%% file: main.tex
\pdfoutput=1

\documentclass{article}

\usepackage{microtype}
\usepackage{graphicx}
\usepackage{subfigure}
\usepackage{booktabs} %
\usepackage{tabularx} %
\usepackage{siunitx} %

\usepackage{xcolor}

\usepackage{listings}
\lstdefinestyle{trace}{ basicstyle=\ttfamily\footnotesize, breaklines=true, 
breakatwhitespace=true, 
columns=fullflexible, 
keepspaces=true, 
frame=single, 
xleftmargin=1em, 
showstringspaces=false }
\usepackage{pifont} 
\definecolor{markyes}{HTML}{009E73}
\definecolor{markpt}{HTML}{E69F00}
\definecolor{markno}{HTML}{D55E00}
\newcommand{\yes}{\textcolor{markyes}{\ding{51}}}
\newcommand{\no}{\textcolor{markno}{\ding{55}}}
\usepackage{tikz}
\newcommand{\pt}{\tikz[baseline=-0.6ex]{\fill[markpt] (0,0) -- (90:0.55ex) arc (90:270:0.55ex) -- cycle; \draw[markpt, line width=0.45pt] (0,0) circle (0.55ex);}}

\usepackage{hyperref}
\hypersetup{hypertexnames=false}
\usepackage{xurl} %

\usepackage[accepted]{mlsys2025}

\mlsystitlerunning{Validation-Driven Deployment of LLM Inference Platforms for Agentic Workloads}
\begin{document}

\twocolumn[
\mlsystitle{Validation-Driven Deployment of LLM Inference Platforms for Agentic Workloads}

\begin{mlsysauthorlist}
\mlsysauthor{Shai Gretz}{to}
\mlsysauthor{Alon Halfon}{to}
\mlsysauthor{Yoav Katz}{to}
\mlsysauthor{Lena Dankin}{to}
\mlsysauthor{Michal Jacovi}{to}
\mlsysauthor{Rachel Tzoref-Brill}{to}
\mlsysauthor{Kfir Toledo}{to}
\mlsysauthor{Elron Bandel}{to}
\mlsysauthor{Natan Ergas}{to}
\mlsysauthor{Michal Malka}{to}
\mlsysauthor{Pravein Govindan Kannan}{to}
\mlsysauthor{Maroon Ayoub}{rh}
\mlsysauthor{Michal Shmueli-Scheuer}{to}
\end{mlsysauthorlist}

\mlsysaffiliation{to}{IBM Research}
\mlsysaffiliation{rh}{Red Hat AI}

\mlsyscorrespondingauthor{Shai Gretz}{avishaig@il.ibm.com}

\mlsyskeywords{Agentic Benchmarking, Agentic Inference, MLSys}

\vskip 0.3in

\begin{abstract}
Enterprises are moving agentic workloads from closed APIs to their own
open-weight model deployments. For this reason, lifecycle-wide validation, from small-scale testing through
sizing and pre-production to production, becomes all the more important. Typically, rollouts are evaluated with ad hoc tools
whose accuracy is rarely checked. We present an end-to-end evaluation
pipeline for agentic LLM serving and empirically ground the design choices that
connect its stages. Our experiments show that replaying captured traces reproduces the per-session
load of the original agent sessions: at the 95th percentile replay stays within
$1.18\times$ of the recorded input tokens while live re-runs reach $4.98\times$,
so a small set of recorded sessions yields a repeatable per-session token load. A code-coverage
analysis of vLLM further shows that the evaluated live, replayed, and synthetic
workloads are complementary in this configuration: each exercises Python serving
paths not reached in the measured runs of the others. In a case study rolling out four large mixture-of-experts (MoE) models on llm-d, experiments show that the target models reached about half the optimal concurrency, and correspondingly about 60\% of the throughput, on captured workloads compared with public traces. We
open-source the relevant tools and release a trace dataset.
\end{abstract}
]

\printAffiliationsAndNotice{}  %

\input{introduction}
\input{workflow_overview}
\input{live_benchmark}
\input{trace_replay}
\input{synthetic_workload}

\input{quality_regression}
\input{vllm_coverage}
\input{usecase}
\input{related_work}
\input{conclusion}

\bibliography{agent_eval}
\bibliographystyle{mlsys2025}

\input{ai_disclosure}

\clearpage 
\onecolumn
\appendix
\input{appendix_dashboard}
\input{appendix_replay}
\input{appendix_synthetic}
\input{appendix_coverage}
\input{appendix_repr}

\end{document}

%% file: introduction.tex
\section{Introduction}
\label{sec:intro}

Agentic workloads are driving rapid growth in LLM inference demand
in enterprises~\citep{goldman2026agentic,aubakirova2026openrouter}.
Coding
assistants, terminal agents, and multi-step workflow automation issue tens to
hundreds of model calls per task, each carrying a context that grows turn by
turn. At this volume, relying on closed-model APIs becomes
expensive~\citep{ey2026agents}, and it also raises concerns about data
governance, vendor lock-in, and control over model versions. As a result, many
organizations are moving to serve open-weight models on their own
clusters~\citep{openrouter2026openweight,maguire2026openweight,tokenomics2026state}, built on serving stacks such as
vLLM~\citep{kwon2023efficient}, SGLang~\citep{zheng2024sglangefficientexecutionstructured}, and
llm-d~\citep{llmd}.

Such an in-house deployment aims to provide an inference service that serves
the organization's agentic workloads at scale: it must support many concurrent
users and agents, keep latency predictable as sessions grow, and match the task
quality users would get from an external provider. Reaching this goal is not a
single step. Every new model, hardware configuration, or serving feature passes
through a sequence of stages before it reaches users, and each stage asks a
different question. Small-scale testing checks that the model runs in the stack
and matches its reference quality. Sizing depends on the shape of agentic
sessions (growing prompts, parallel tool calls, sub-agent fan-out, and idle
gaps while tools run), so it needs realistic load replayed at rising
concurrency. Qualifying a release requires real agents solving real tasks, to
show that the platform serves them as well as an external provider does. In
production, the question becomes what changed between releases, which requires
telling real regressions apart from the run-to-run variance of
non-deterministic agents.

Evaluating across these stages faces two gaps. \emph{First, the stages are
disconnected.} Agent evaluation
frameworks~\citep{shi2026harbor,inspectai,bandel2026general} measure agent
quality but treat the inference endpoint as fixed, while load
generators~\citep{aiperf2026,chandrasekar2026inferenceperf,guidellm2024} measure serving
performance using synthetic prompts or replayed traces. Recent work such as XPerf~\cite{wang2026xperf} demonstrates that traces from real agentic applications can be collected and reproducibly replayed across LLM serving systems. However, XPerf focuses on benchmarking and profiling serving systems, whereas our goal is to provide a connected evaluation pipeline spanning agent development, deployment, and serving evaluation. Public
traces~\citep{semianalysis_weka_060826,qin2024mooncake} drop the request content,
so they cannot be re-issued against the target model. Because these tools rely
on different data, what one stage captures cannot be fed directly to the next.
\emph{Second, the evaluation choices are seldom checked.} Every stage depends
on which workloads to run, which traces to replay, and which tool to replay
them with. Teams deploying models follow similar rollout steps on their own
platforms~\citep{baseten2026kimik3,google2026llmd}, but make these choices by
experience, without measuring whether a workload exercises the serving paths
agents use, whether traces represent the target traffic, or whether replay
reproduces the recorded sessions. A poor choice gives a misleading answer:
sizing a deployment on traces unlike its actual traffic yields capacity
estimates that do not hold in production.

We address both gaps. \emph{To connect the stages}, we design live
benchmarking to record every model call in full, so that the runs used to
measure quality also produce the traces needed for sizing, and build the
pipeline around this shared replayable format. \emph{To ground the design},
we empirically examine the choices that connect these stages and let the results
shape the design. We then apply the pipeline while rolling out large mixture-of-experts (MoE)
models on llm-d. We focus on performance and quality; functional API tests, which check that
the deployed serving API handles each type of request correctly, serve as a
first gate, and security or safety
evaluation is out of scope.

\paragraph{Contributions.} This paper makes the following contributions: (1)~\textbf{A connected lifecycle evaluation pipeline for agentic LLM serving}
that covers quality and performance from small-scale testing to production,
combining live benchmarking, dependency-aware trace replay, synthetic workload
generation, and regression analysis over a single trace format;
(2)~\textbf{Empirical grounding of cross-stage design choices}: vLLM statement coverage
shows that the evaluated live, replayed, and synthetic workloads are complementary
in this configuration: each reaches Python serving paths absent from the measured
runs of the other families; an analysis
of existing trace formats and replay systems identifies what agentic replay
requires; and replay reproduces the per-session load of recorded agent
sessions: it
keeps the 95th-percentile input within $1.18\times$ of the recording versus up
to $4.98\times$ for live re-runs, with stable load across repeated runs;
(3)~\textbf{Lessons from real-world rollouts} of four MoE models on H100 and H200 GPUs, showing how workloads and tools were used across stages. Public traces sufficed for initial configuration, but traces captured on all target models reached roughly half the optimal concurrency and 60\% of the throughput, and exposed a routing imbalance before deployment;
(4)~\textbf{Open-source tools and data}: Workflow-level OpenTelemetry trace replay, per-session error reporting, and synthetic agentic workload generation were contributed upstream to Inference Perf~\cite{chandrasekar2026inferenceperf}. We also extended Exgentic with container-native live benchmarking and release a dataset containing 10,056 agent sessions with full request contents.

%% file: workflow_overview.tex
\section{Workflow Overview}
\label{sec:workflow_overview}

When deploying a new version of a production agentic system, teams must quantify the impact of a broad range of configurations, including modifications to KV-cache mechanisms, hardware configurations (e.g., additional GPUs), model updates, and agent behavior. We first describe the stages a deployment goes through on its way to production and the questions teams need to answer at each stage (Section~\ref{sec:dev_stages}). We then derive from these questions the evaluation capabilities required for the development process, and map each capability to the respective component of our framework (Section~\ref{sec:dev_needs}).

\subsection{Stages of Development}
\label{sec:dev_stages}

\begin{figure*}[t]
  \centering
  \includegraphics[width=\textwidth]{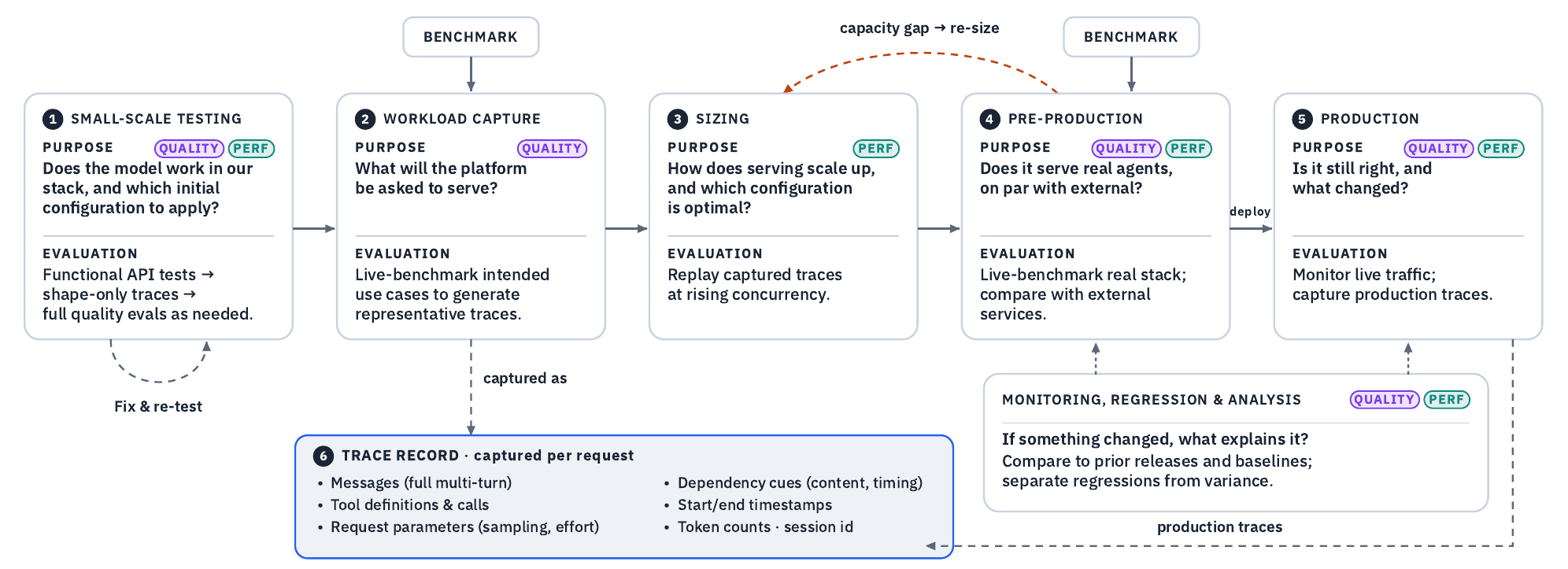}
  \caption{Integrated evaluation along the path to deployment.}
  \label{fig:e2e_workflow}
\end{figure*}

We view the path to deployment as a sequence of five lifecycle stages (Figure~\ref{fig:e2e_workflow}). Each stage is defined by its \emph{purpose}, a question about the system, and by the \emph{evaluation} that answers it:

\paragraph{Stage 1: Small-scale testing.} The model is deployed on a few GPUs and checked in increasing order of cost before any scale-out. \emph{Functional API tests} come first. They send short, scripted requests through the deployed serving stack and check each response against known expected behavior, such as chat completion, streaming, tool calling, reasoning output, structured output, and clean errors for malformed requests. At this stage integration errors are caught cheaply, before any agent runs. Shape-only traces then drive initial sizing and tuning. Finally, real agents run on representative benchmarks to confirm that the model matches its reference quality on the intended use cases. This stage answers: \emph{Does the model and initial configuration work in our stack?}

\paragraph{Stage 2: Workload capture.} Before the platform can be sized or qualified, we must know what it will serve. Given the intended use cases, we target relevant open-source benchmarks and run them with real agents on the target model, producing a set of representative sessions. These sessions retain the full content, including inputs, outputs, and tool definitions (the trace record, item~\#6 in Figure~\ref{fig:e2e_workflow}). Running real agents at this point also surfaces gross quality problems, such as a model that cannot drive a given agent scaffold, long before the release is qualified. This stage answers: \emph{What will the platform be asked to serve?}
\paragraph{Stage 3: Sizing.} Once quality is established, the deployment grows to its target scale using the captured workload. The required hardware depends on the \emph{session} shape of agentic traffic rather than on individual requests: prompts that grow turn by turn, sub-agent fan-out, and idle gaps while tools execute. The recorded traces are replayed at increasing load to size the deployment and find where throughput saturates. This stage answers: \emph{How does serving scale up, and which configuration is optimal?}

\paragraph{Stage 4: Pre-production.} This stage is the acceptance gate. The candidate version is deployed end to end on the release images, with real agents, models, and tools, and is served with the real benchmark workloads. Agent success is compared with the previous version as well as external results. Pre-production also checks behavior under contention, such as priority and fairness between agent traffic and interactive users. This stage answers: \emph{Does it serve real agents on par with external baselines?} A capacity gap found here loops back to sizing.
\paragraph{Stage 5: Production.} After deployment, live traffic is itself the benchmark, since no trace fully matches it. Latency and throughput are tracked per tenant at the router and at the engines, and each release is compared with its predecessor. Production usage also keeps evolving: new agents, tools, and user behaviors appear, and a version validated against yesterday's workload may not hold up against today's. This stage answers: \emph{Is it still right, and what changed?} Degradation at this stage feeds back to pre-production and sizing.
\paragraph{Monitoring, Regression, and Analysis.} Analysis and regression tools can be used to identify performance degradations during pre-production validation and production benchmarking. When a metric deviates from a previous release or an external baseline, the root cause must be isolated, whether it originates in the model, agent framework, serving infrastructure, or workload itself, enabling the change to be corrected, rolled back, or intentionally accepted. Because live agent executions are inherently non-deterministic, effective regression analysis must also distinguish true regressions from normal run-to-run variability. This capability answers: \emph{If something changed, what explains the observed behavior?}

\subsection{From Development Stages to Evaluation Needs}
\label{sec:dev_needs}

Answering the questions above requires a set of evaluation capabilities, which we now derive from the stages in Figure~\ref{fig:e2e_workflow} and map to the components of our framework.

\paragraph{Live end-to-end benchmarking.} Workload capture and pre-production both require running real agents, models, and benchmarks against a deployment, since synthetic prompt loops fail to capture the complexity and diversity of real-world agent interactions. During workload capture, live benchmarking characterizes the intended use cases and checks agent success against publicly available results; during pre-production, it monitors task success, latency, throughput, and resource utilization on the candidate stack in addition to comparing them against external baselines and prior releases (Section~\ref{sec:live-benchmark}).

\paragraph{Full-content replayable traces.} Collecting and repeatedly executing live workloads is costly and operationally challenging. Live benchmarking therefore records each request as a \emph{trace} in a replayable format. Each record preserves the full multi-turn messages, tool definitions and calls, request parameters (e.g., sampling settings and reasoning effort), timestamps, token counts, and session identifiers. Inter-request dependencies are not explicit; replay infers them from message content and timing, then derives post-predecessor delays. In contrast, existing traces such as WEKA \cite{semianalysis_weka_062126} hash away the input and are therefore insufficient for robust sizing. This is what makes faithful replay possible. Production traffic can be recorded in the same format, so once its capture is permitted, production traces can flow back into the workflow without conversion (Section~\ref{sec:trace_format}).\footnote{Capturing the full content of production traffic generally requires explicit user consent, and is left to future work.}

\paragraph{High-fidelity trace replay.} Sizing requires reconstructing and executing the recorded workload against the serving stack at controlled, increasing load, while scheduling requests according to inferred dependencies and delays derived from recorded timestamps. Replaying traces isolates the serving stack from agent and tool variability and lets the same workload be re-run across hardware configurations and releases (Section~\ref{sec:trace-otel-replay}).

\paragraph{Synthetic agentic workload generation.} Synthetic workloads cover agentic execution patterns that are rare or absent in real traces, such as hierarchical agent spawning, latency-intensive tool calls, and long-running user-agent dialogues, to further facilitate configuration tuning in the sizing stage (Section~\ref{sec:synthetic}). 

\paragraph{Monitoring, analysis, and regression detection.} The analysis and regression capability requires tools that track runs across releases, compare them, detect regressions, and explain them. In pre-production, these tools decide whether the candidate passes the acceptance gate by comparing it against prior releases and external baselines; in production, they compare per-release metrics of live traffic and flag drift as usage evolves. We combine rule-based checks with agent-assisted analysis to investigate root causes (Section~\ref{sec:dashboard}).

While existing approaches for live benchmarking (e.g., Harbor~\citep{shi2026harbor}) and trace replay (e.g., AIPerf~\citep{aiperf2026}) address parts of this process, they typically provide only partial coverage of the evaluation lifecycle. In contrast, our framework combines synthetic workloads, live benchmarking, trace replay, and analysis into a unified end-to-end pipeline, as described in the following sections.

%% file: live_benchmark.tex
\section{Live Benchmarking}
\label{sec:live-benchmark}

Unlike synthetic data generation or trace replay, live benchmarking directly measures end-to-end task quality, the only component of our pipeline to do so, making it the best proxy for the quality of agentic executions.

Hundreds of agentic benchmarks now exist, spanning software engineering
(e.g., DeepSWE~\cite{huang2026deepswe}), hardware design
(e.g., HWE-Bench~\cite{cui2026hwebench}), computer and terminal use
(e.g., Terminal-Bench~\cite{terminalbench}), office and cross-application
workflows (e.g., AutomationBench~\cite{shepard2026automationbench}),
and mathematics (e.g., FrontierMath~\cite{glazer2024frontiermath}).
Running them individually does not scale: each demands its own setup
expertise and operational care. Frameworks such as
Harbor~\cite{shi2026harbor}, Inspect AI~\cite{inspectai},
and Exgentic~\cite{bandel2026general} address this by consolidating
multiple benchmarks behind a common interface, but consolidation alone does not suffice for agentic platform evaluation. 

\textbf{Coverage.} The harness must run public benchmarks representative
of expected deployment use cases, and allow adding proprietary benchmarks with
modest development effort. Because benchmarks, agent scaffolds, and models each shape the traffic reaching the inference endpoints, and use cases may mandate particular models or scaffolds, the harness must also support all relevant combinations across the three axes.   

\textbf{Scale.} Evaluation must execute many benchmarks concurrently as agentic tasks can take hours to complete. For example, Long-Horizon-Terminal-Bench 2.0 \cite{li2026longhorizonterminalbench20automatingdiverse} assigns tasks a three-hour budget and reports that 21\% of solved tasks take over two hours. Execution must therefore be robust and consistent in local and cloud environments, handling transient failures and retries.

\textbf{Trace recording fidelity.} Concurrent execution is eventually bounded by the resources each task needs to stand up its benchmark and agent; some environments span multiple containers or virtual machines, consuming many gigabytes of memory and storage per task. To scale further, LLM call traces must be recorded at high fidelity once and then replayed concurrently at much larger scale.

\textbf{Reporting.} Each evaluation must report three distinct levels of measurement: task quality
(benchmark success), task-level performance (end-to-end latency, token
consumption, and mean per-call prefix-cache reuse, measured as the fraction of
input tokens reported as cached), and per-call behavior (latency, output rate,
LLM errors).

We address these requirements by extending Exgentic, an open-source framework for capturing, storing, analyzing, and recording agentic sessions, along three lines.

First, we adopt a fully container-native design\footnote{\url{https://github.com/Exgentic/eval-containers}} in which an evaluation is represented as the composition of four independent dimensions---benchmark, agent, model, and task. Each benchmark is packaged as a self-contained Open Container Initiative (OCI)-compliant image, so evaluations run directly under Docker Compose or Kubernetes without a dedicated evaluation framework at runtime. The resulting architecture separates experimental methodology from runtime orchestration, making large-scale evaluations easier to reproduce across local and cloud deployments.

Second, every model executes behind a logging
gateway that records each LLM call in full fidelity at the HTTP transport level. Inference Perf replays these records directly or as OpenTelemetry (OTel) GenAI spans (Section~\ref{sec:trace-replay}). Each run thus stores the full messages, tool calls, latencies, and token usage alongside benchmark scores for later replay and analysis, improving auditability, debugging, and reproducibility.

Finally, a dashboard UI launches evaluations and displays their results, highlighting inference-platform issues such as LLM errors and latency alongside standard benchmark quality metrics (Section~\ref{sec:dashboard}).

%% file: trace_replay.tex
\section{Trace Replay}
\label{sec:trace-replay}

\input{trace_format}

\subsection{Workflow-Level OTel Replay}
\label{sec:trace-otel-replay}

These full-content traces drive workflow-level replay of agentic patterns involving tools, parallel branches, and multiple steps. Request-level replay reproduces arrival patterns and token lengths, conversation replay preserves sequential history, and graph-capable replay additionally schedules concurrent calls from a dependency graph. We contributed trace replay\footnote{\url{https://github.com/kubernetes-sigs/inference-perf/blob/main/docs/otel_trace_replay.md}} to Inference Perf: it reads agent traces exported in the OpenTelemetry (OTel) GenAI format~\citep{opentelemetry-specification}, reconstructs each workflow, and executes it against the target endpoint.

The replay pipeline comprises trace ingestion, inferred dependency-graph construction, dependency-aware execution, and live-output propagation.

Rather than replaying requests from timestamps alone, the framework infers predecessor--successor relationships among LLM invocations. It adds causal edges when a request contains an assistant message matching an earlier span's output, uses the closest non-overlapping earlier span as a temporal fallback, and applies transitive reduction. Roots run immediately; other calls wait for every predecessor and then observe their derived post-predecessor delay. The scheduler supports sequential chains, parallel branches, and fan-out/fan-in synchronization in the inferred graph.

Replay is also \emph{output-aware}: live generated text and structured tool calls are substituted into dependent requests, rather than only recorded responses. To preserve tool-call fidelity, a turn with exactly one recorded function available in its tool definitions forces that function; turns with multiple or unmatched recorded functions require a tool call without selecting a specific function. This keeps live generation structurally aligned with the recorded tool-result messages reused downstream, without executing external tools.

Because a failed request can affect or cancel downstream requests, the per-load-condition error report we contributed to Inference Perf records request- and session-level failures (Appendix~\ref{app:replay}).

\paragraph{Comparison with other replay systems.} Table~\ref{tab:replay_capabilities} in Appendix~\ref{app:replay} compares Inference Perf with five other replay systems; the broader evaluation workflow is replay-system agnostic.

\subsection{Evaluation}
\begin{table}[t]
  \centering
  \footnotesize
  \setlength{\tabcolsep}{3pt}
  \caption{Per-session load ratios relative to the recorded traces (replayed or re-run total $\div$ recorded total), median and P95 over 40 sessions per benchmark. Replay: replaying workload in Inference Perf; Re-run: reproducing live benchmark in Exgentic. DeepSeek stands for DeepSeek-V4.1-Flash-1017.}
  \label{tab:replay_load_ratio}
  \resizebox{\columnwidth}{!}{%
  \begin{tabular}{llcccccc}
    \toprule
    & & \multicolumn{2}{c}{$R_{\mathrm{in}}$} & \multicolumn{2}{c}{$R_{\mathrm{out}}$} & \multicolumn{2}{c}{$R_{\mathrm{peak}}$} \\
    \cmidrule(lr){3-4} \cmidrule(lr){5-6} \cmidrule(lr){7-8}
    \textbf{Type} & \textbf{Benchmark (Model)} & Med. & P95 & Med. & P95 & Med. & P95 \\ \midrule
    Replay & HWE-Bench (GLM-5.3) & 0.91 & 1.12 & 0.74 & 1.28 & 0.89 & 1.14 \\
    Re-run & HWE-Bench (GLM-5.3) & 1.01 & 4.98 & 1.09 & 3.84 & 1.01 & 1.85 \\ \midrule
    Replay & CyberGym (DeepSeek) & 1.00 & 1.18 & 1.05 & 2.04 & 1.02 & 1.30 \\
    Re-run & CyberGym (DeepSeek) & 1.14 & 3.97 & 1.16 & 4.52 & 1.06 & 1.76 \\
    \bottomrule
  \end{tabular}}
\end{table}
We evaluate how closely replay matches the recorded session's load and call/tool sequences, and how consistent repeated replays are (metric definitions and setup in Appendix~\ref{app:replay_eval}).

\paragraph{Fidelity.} We compute per-session ratios of replayed to recorded input tokens ($R_{\mathrm{in}}$), output tokens ($R_{\mathrm{out}}$), and peak context, i.e., maximum input size ($R_{\mathrm{peak}}$). We also measure the normalized edit similarity of call-type sequences (\emph{workflow fidelity}, WF) and the per-turn overlap of invoked tool names (\emph{tool-call fidelity}, TF).

We collect traces with Exgentic using the Claude Code~\cite{anthropic2025claudecode} agent harness, by running GLM-5.3~\cite{glm5team2026glm5vibecodingagentic} on 40 HWE-Bench tasks and DeepSeek-V4.1-Flash-1017~\cite{deepseekai2026deepseekv41flash} on 40 CyberGym~\cite{wang2026cybergym} tasks. We replay each trace with Inference Perf using its default tool-choice forcing policy. As a reference, not a competing method, we re-run the same tasks live in Exgentic; this shows how far a fresh sample of the agent drifts from its recording, the run-to-run variance that replay is meant to remove. 

Table~\ref{tab:replay_load_ratio} presents the results. At the median, replay and live re-runs are similar: $0.74$--$1.05\times$ for replay and $1.01$--$1.16\times$ for re-runs. The tails differ sharply. Replay keeps the P95 $R_{\mathrm{in}}$ within $1.18\times$ and the P95 $R_{\mathrm{peak}}$ within $1.30\times$, whereas re-runs reach $4.98\times$ and $1.85\times$. The P95 $R_{\mathrm{out}}$ is $1.28$--$2.04\times$ for replay and $3.84$--$4.52\times$ for re-runs. Replay thus bounds per-session load, partly by design because it reuses recorded tool outputs and enforces tool selection, demonstrating that propagating live outputs over long sessions does not substantially erode this property compared to live re-runs. Replay also closely matches the recorded call-type and tool-name sequences: WF is $0.92$--$0.94$ for replay vs.\ $0.68$--$0.70$ for re-runs, and TF is $0.87$--$0.90$ vs.\ $0.20$--$0.30$, as expected due to tool-call enforcement.

\paragraph{Consistency.} We replay one trace per benchmark 50 times and compare the replays with each other. We report the \emph{coefficient of variation (CV) of per-replay time per output token (TPOT)} and \emph{workflow consistency (WC)}, which is WF averaged over all replay pairs. WC is computed on call types only ($\mathrm{WC}_{\mathrm{role}}$) and with each assistant step labeled by the tools it called ($\mathrm{WC}_{\mathrm{tool}}$). Per-replay TPOT is token-weighted, $\mathrm{TPOT}_i = \sum_r d_r / \sum_r (o_r - 1)$, where $d_r$ is the decode time of request $r$ (latency minus time to first token) and $o_r$ is its output token count, so short calls do not dominate.

Table~\ref{tab:replay_consistency} shows a TPOT CV of 5.2\%--10.7\%, so replay places a stable load on the server. Replays agree almost completely at the role level. For GLM-5.3 on HWE-Bench, agreement drops once tool choices are included: live sampling keeps the workflow shape but sometimes picks a different tool at the same step.

\begin{table}[t]
  \centering
  \footnotesize
  \setlength{\tabcolsep}{4pt}
  \caption{Inter-replay consistency over 50 replays of one trace: CV of per-replay TPOT, and workflow consistency with call-type symbols only ($\mathrm{WC}_{\mathrm{role}}$) and with tool-labeled assistant steps ($\mathrm{WC}_{\mathrm{tool}}$), averaged over all replay pairs. DeepSeek stands for DeepSeek-V4.1-Flash-1017.}
  \label{tab:replay_consistency}
  \begin{tabular}{llccc}
    \toprule
    \textbf{Benchmark (Model)} & \textbf{TPOT CV} & $\mathbf{WC_{role}}$ & $\mathbf{WC_{tool}}$ \\ \midrule
    HWE-Bench (GLM-5.3) & 5.2\% & 92\% & 86.8\% \\
    CyberGym (DeepSeek) & 10.7\% & 99.6\% & 99.3\% \\
     \bottomrule
  \end{tabular}
\end{table}

%% file: trace_format.tex
\subsection{Replayable Trace Formats}
\label{sec:trace_format}

The traces captured during live benchmarking are consumed directly by replay. They record semantic request content (messages, tool calls and definitions, and request parameters) and serving metadata (token counts and timing). Because full request content is retained, replay can reconstruct shared prefixes and thereby induce prefix reuse, although the traces contain no explicit prefix-sharing or cache signal. Replay infers dependencies from content and timing, while agent hierarchy remains unannotated. Existing formats emphasize either agent behavior, as in SWE-smith~\cite{yang2025swesmithscalingdatasoftware} and TraceLab~\cite{zhu2026tracelabcharacterizingcodingagent}, or workload shape and cache behavior, as in WEKA~\cite{semianalysis_weka_062126}, Mooncake~\cite{qin2024mooncake,qin2025mooncake}, and Baseten~\cite{aiperf_baseten_trace}; none retains the structured content needed to re-issue agent requests verbatim.

Exgentic traces fill this gap: each LLM call is an OTel GenAI chat span holding the complete input and output messages, tool definitions, request parameters, token counts, and timestamps, so requests can be re-issued without synthesizing their content. We release 10,056 such sessions (241,674 LLM calls), spanning six benchmarks, five models, and five agent harnesses~\cite{exgentic-agent-llm-traces-v2}. Appendix~\ref{app:replay} compares the formats (Table~\ref{tab:trace_formats}), details the Exgentic schema, and shows an example session (Figure~\ref{fig:exgentic_example}).

%% file: synthetic_workload.tex
\section{Synthetic Agentic Workload Generation}
\label{sec:synthetic}

Trace-driven replay is limited to traces that have already been collected. We therefore contributed \emph{synthetic agentic workload generation} to Inference Perf,\footnote{\url{https://github.com/kubernetes-sigs/inference-perf/blob/main/docs/synthetic_agentic.md}} which procedurally constructs agentic sessions from a few configurable parameters to stress-test dimensions of agent behavior that are hard to isolate in recorded traces: deep tool-use loops, wide parallel tool calls, recursive sub-agent delegation, long-context accumulation, and context compaction. The generator emits the same \textit{ReplayGraph} used by trace replay, so synthetic and trace-driven runs share one execution and measurement path. Token counts, loop depth, fan-out, and latencies follow configured distributions, a declarative \emph{theme} supplies semantically coherent domain content so the model under test responds plausibly, and all stochastic choices derive from a seed for reproducibility (more details in Appendix~\ref{app:synthetic}).

%% file: quality_regression.tex
\section{Monitoring, Regression, and Analysis}
\label{sec:dashboard}

Our dashboard launches, monitors, and analyzes Exgentic live benchmarking jobs. Users select a benchmark$\times$agent pair, a model, and an inference endpoint, then follow each job's progress, pass rate, token usage, and cost (Figure~\ref{fig:dashboard_runs}); each task's downloadable trace is the same artifact consumed by trace replay (Section~\ref{sec:trace-replay}), so any inspected run can be replayed at scale (Figures~\ref{fig:dashboard_run_detail}--\ref{fig:dashboard_task_calls}). To assess deployment changes, regression suites, seeded automatically per model with fast, previously passing tasks, compare two executions down to the tasks that regressed from pass to fail (Figure~\ref{fig:dashboard_regression}). On request, a deterministic pass computes a run's factual figures, Claude Code interprets them, and a validator rejects any report whose facts differ from the precomputed values (Figures~\ref{fig:dashboard_report_metrics} and~\ref{fig:dashboard_report_trajectories}). Screenshots and further details are given in Appendix~\ref{app:dashboard}.

%% file: vllm_coverage.tex
\section{Workload Complementarity via Code Coverage}
\label{sec:vllm_coverage}

The evaluation pipeline combines multiple approaches, each intended to capture behaviors the others miss. We quantify this by instrumenting a real vLLM deployment, running each approach as a separate workload, and measuring vLLM source coverage. Workloads that execute python statements no other workload reaches are considered complementary.

\subsection{Experimental setup}

We deploy vLLM~\citep{kwon2023efficient} \texttt{v0.31.0} serving
DeepSeek-V4.1-Flash-1017 on an eight-H200-GPU node, in an image modified to
collect Python statement coverage from the live server with
coverage.py~\citep{coveragepy}. Coverage is attributed to individual runs
without redeploying the server. Appendix~\ref{app:coverage-method} gives the
server command line (Figure~\ref{fig:serve-cmdline}) and the isolation
mechanism.

The installation has 419{,}671 measurable Python statements across 2{,}705
files. Much of this code is unreachable by construction: one vLLM install
ships model implementations, hardware backends, and optional features that a
given deployment never loads. Absolute coverage is therefore not meaningful on
its own; we report each workload's \emph{relative} contribution. To make
absolute figures interpretable, a static reachability analysis seeded from the
modules the server command line selects rules out 180{,}352 statements
(43.0\%) this deployment cannot import, and we report coverage against the
239{,}319 that remain
(Appendix~\ref{app:coverage}, Table~\ref{tab:cumulative-coverage}).

\subsection{Workloads}

We measure the statement coverage of vLLM serving DeepSeek-V4.1-Flash-1017
under five test families: \emph{Trace replay} is represented by WEKA traces~\citep{semianalysis_weka_060826}
replayed with AIPerf and Exgentic agent
traces replayed with Inference
Perf, kept as two families because the
corpora differ in content and in the tool driving them;
\emph{synthetic agentic workloads} are generated with Inference Perf; and
\emph{live benchmarking} runs CyberGym agents with
Exgentic. The \texttt{smoke} family is a single non-streaming request, included
as a cold-start reference point rather than as a pipeline stage.

\subsection{Results}

Table~\ref{tab:unique-coverage-main} shows what each family covered and the
statements it alone reached. Appendix~\ref{app:coverage} gives per-run figures
(Table~\ref{tab:coverage-workloads}), reachable-statement coverage
(Table~\ref{tab:cumulative-coverage}), expanded unique coverage
(Table~\ref{tab:unique_coverage}), and the hard-to-reach scheduling, caching
and KV-offload states each workload induced (Table~\ref{tab:load-states}).

\begin{table}[t]
\centering
\footnotesize
\caption{Coverage by test family, request-driven only (server startup
excluded). \emph{Cov.} is the union of the statements the family's runs
executed in their own isolated windows; \emph{Uniq.} is the statements in that
union that no \emph{other} family reached.}
\label{tab:unique-coverage-main}
\setlength{\tabcolsep}{4pt}
\begin{tabular}{@{}p{0.155\columnwidth}@{\hspace{4pt}}rrrp{0.37\columnwidth}@{}}
\toprule
\textbf{Family} & \textbf{Runs} & \textbf{Cov.} & \textbf{Uniq.} & \textbf{Reached only by this family} \\
\midrule
live & 1 & 8{,}777 & \textbf{331} & Anthropic messages path; aborted KV transfer \\
weka & 3 & 8{,}431 & 54 & streaming tool-call parsing; deferred offload fences \\
exgentic-\allowbreak traces & 3 & 8{,}967 & 27 & sparse-attention short-context branch \\
smoke & 1 & 5{,}748 & 20 & greedy sampling \\
synthetic & 16 & 8{,}795 & 5 & sparse-SWA chunk cutoff \\
\midrule
\textbf{Union} & \textbf{24} & \textbf{9{,}778} & \textbf{437} & \\
\bottomrule
\end{tabular}
\end{table}

Including server startup, the \emph{server-wide cumulative} total of the 24
runs is 74{,}562 of the 239{,}319 reachable statements (31.16\%), dominated by
the 69{,}127 executed while loading the model, initializing KV caches, and
warming up kernels. In each run's \emph{own isolated coverage window}, the
request-driven union is 9{,}778 statements.

\paragraph{API Surface Coverage.} A workload's value lies in which code it reaches, not how much. Most unique statements reflect API and client choices rather than differences in engine behavior. The \texttt{live} family contributes 331 unique statements largely because it alone exercises the Anthropic Messages API; most of \texttt{weka}'s 54 and \texttt{smoke}'s 20 stem from their unique use of streaming and temperature-0 sampling, respectively. Because statements reached by more than one family are not counted as unique, these totals provide only a lower bound on how much the families differ.

The API surface is nonetheless a real thing to test, not an artifact of our
setup: a deployment serves whichever endpoints its clients use, and in our experience
(Section~\ref{sec:usecase}) a mismatched tool parser can degrade quality on one
path while leaving others healthy, while a LiteLLM~\citep{litellm} gateway may
convert between the OpenAI and Anthropic formats in ways that introduce errors
exposed only on specific APIs. A suite covering one endpoint would miss both.
The remainder of the column is not API surface at all: the
\texttt{exgentic-traces} and \texttt{synthetic} entries involve no endpoint or
client setting the other families lacked.

\paragraph{Load-Driven Paths.} GPU-to-CPU offloading code was touched by every family, but at very different
depths. The single-request \texttt{smoke} baseline reached only 4 statements of
the CPU offload worker---worker setup, with no block ever transferred---and none
of the eviction policy, whereas \texttt{live} reaches 112 statements of the
offload worker and 26 of the eviction policy, and the \texttt{weka} replays 112
and 28 respectively.

\paragraph{Family-Specific Behaviors.} Within those shared paths, individual families still reach distinct code. In
\texttt{weka}, a request completed before its KV-cache copy to CPU had
finished, leaving a deferred fence that must be reconciled when the block is
reused; this is the one KV-offload path no other family reaches
(Table~\ref{tab:load-states}). In \texttt{live}, a session was
terminated, causing an abort of the in-flight KV-cache transfer. The
\texttt{exgentic-traces} replays were the only family to enter the
short-context branch of sparse attention, taken when a request offers fewer
index candidates than the top-$k$ width.

\paragraph{Closing a Coverage Hole with a Synthetic Workload.} While
GPU-to-CPU offloading was entered by every sustained family, the reverse
CPU-to-GPU \emph{restore} path was executed by none of the 24 runs, exposing a
real coverage hole. It requires a working set exceeding the GPU KV pool
\emph{and} revisits spaced far enough apart to evict a prefix first---a
conjunction no workload in our set satisfied, but one a synthetic workload sets
directly. This covered 124 new statements of the offload and
tiering subsystem, demonstrating the value of synthetic generation
(Appendix~\ref{app:coverage-kvrestore}).

\subsection{How much testing is enough?}

Figure~\ref{fig:coverage-growth} plots the cumulative union of request-driven
statements against cumulative testing time, per family and overall, with each
series ordered shortest-run-first so the horizontal axis is the testing budget
spent. Saturation is evident. The \texttt{synthetic} family reaches 8{,}581
statements after 2 cumulative minutes and 8{,}795 after 240.5: its two
longest runs occupy 151 of those minutes and add 22 statements between them.
The 1-hour \texttt{weka} run adds 33 statements over the 20-minute one at
the same concurrency, and loses 32 others, so tripling the load time leaves
coverage essentially unchanged.
 Long runs still measure sustained performance, but coverage stops
distinguishing them early, so a coverage-guided suite would spend its minutes
on more workloads rather than longer ones.

\subsection{Limitations}

Similar statement coverage does not show that tests are redundant. Coverage
collection is incomplete, since only vLLM's Python code is measured, not its
compiled kernels; a covered line was executed but not shown to be correct, and
the approaches differ in what they can check; and the results come from a single
model, hardware, vLLM configuration, and selected set of workloads (details in
Appendix~\ref{app:coverage-limits}).

%% file: usecase.tex
\section{Case Study: Bringing Up MoE Models on llm-d}
\label{sec:usecase}

We now show how the llm-d team~\cite{llmd} uses the workflow of Section~\ref{sec:workflow_overview} and our tools in practice. llm-d is an open-source, Kubernetes-native distributed inference platform that runs on model servers such as vLLM and SGLang. It serves open-weight models at scale on heterogeneous hardware, adding an inference-aware router (the endpoint picker, EPP), KV-cache offloading, prefill/decode (P/D) disaggregation, and wide expert parallelism. Each feature must be tuned per model, model server, and accelerator, and its value depends on the shape of the served workload.

We brought up four large MoE models: GLM-5.3~\cite{glm5team2026glm5vibecodingagentic}, Nemotron 3 Ultra~\cite{nvidia2026nemotron3ultraopen}, DeepSeek-V4.1-Flash-1017~\cite{deepseekai2026deepseekv41flash},\footnote{Initially, some experiments were made with DeepSeek-V4.1-Flash before moving to DeepSeek-V4.1-Flash-1017.} and Kimi-K3~\cite{kimiteam2026kimik3openfrontier}, on H100 and H200 GPUs, following the stages of Figure~\ref{fig:e2e_workflow}, with the sizing stage run twice: first for the model servers, then for the router.

\paragraph{Small-scale testing.} Functional API tests confirmed that chat completion, tool calling, and streaming work through the llm-d stack, and sizing arithmetic chose tensor and expert parallelism, GPU count, and KV-cache memory. With no captured workload yet, we tuned against WEKA traces~\cite{semianalysis_weka_062126}\footnote{These WEKA traces are a later release of the same corpus~\cite{semianalysis_weka_060826} used in Section~\ref{sec:vllm_coverage} and Appendix~\ref{app:repr}; those experiments use the release that was available when they were set up.} on H100 and NVIDIA Moontrace traces~\cite{nvidia_moontrace_2026} on H200, replaying both with AIPerf. 
We replayed traces at eight increasing concurrency levels, restarting the server for a cold cache, and ran each configuration until performance degraded; these sweeps identified the operating point and guided tuning of prefix caching, CPU KV offloading, multi-token prediction (MTP), and P/D disaggregation.

\paragraph{Workload capture.} Public traces describe someone else's workload, and WEKA hashes away message content, so its requests cannot be re-issued against the target model (Section~\ref{sec:trace_format}). We therefore selected benchmarks spanning the model's potential uses, e.g., Terminal-Bench~2.1~\cite{terminalbench},\footnote{\url{https://github.com/harbor-framework/terminal-bench-2-1}} HWE-Bench, and SWE-Bench Multimodal (MM)~\cite{yang2025swebench}, which adds image inputs. We ran them with real agents on the deployed model using Exgentic (Section~\ref{sec:live-benchmark}), producing full-content traces.

\paragraph{Sizing, step 1: model servers.} We replayed the captured traces with the OTel trace replay of Inference Perf (Section~\ref{sec:trace-replay}), and used session duplication with cache-isolating markers to reach high concurrency from a limited set of recorded sessions. Compared with the public traces used for tuning, the captured workloads reached around half the optimal concurrency (the highest load level before performance degraded) for all target models, and per-GPU throughput was correspondingly lower, around 60\% of that achieved with the public traces. We attribute this gap mainly to differences in the workloads themselves: completion lengths and session shapes depend on the model that generated the trace, and the public traces were not recorded on the target model, whereas the captured traces were. 

\paragraph{Pre-production $\rightarrow$ sizing, step 2: router.} With the model servers sized, we repeated the sizing stage across replicas to configure the router. The router balances a KV-cache prefix-affinity scorer, which keeps a session on the replica holding its KV cache, against a token-load scorer, which avoids overloading that replica. Flow control adds priority and fairness under contention: a latency-sensitive workload with a TTFT target receives higher priority and reserved capacity, while standard and batch traffic are capped at lower saturation ceilings. We first tuned the scorers by replaying WEKA traces with AIPerf at rising load across replica counts, and tuned flow control with fixed prefill-heavy and decode-heavy workloads, accepting a configuration only if the high-priority workload met its TTFT target.

We then replayed the captured agentic workloads at scale with Inference Perf. Terminal-Bench~2.1 sessions all begin with a large identical prompt prefix, and under this workload the router sent most traffic to the first few replicas that had cached that prefix, leaving the others almost idle: prefix affinity dominated because the token-load scorer never compensated for the cache reuse. We fixed the imbalance by switching the token producer feeding the token-load scorer from estimate to accurate token computation (real tokenization). The bug appeared only under a workload with the structure of real agent traffic, which production traffic is also expected to have: every session in these traces begins with the same tool definitions (about 17K tokens), followed by a system prompt of about 26K--27K characters that differs only slightly between sessions. In contrast, the WEKA traces scope prefix hashes to each session, so replayed sessions share no common prefix, and their first substantive requests vary widely (29K--98K tokens, p5--p95).\footnote{The final engine configuration is proprietary and is not shared.}

\paragraph{Production: workload representativeness.} After deployment, we checked
whether captured workloads represent what the platform serves, by running an experiment comparing
per-request input and output token-length distributions of typical agentic workload on the
GLM-5.3 deployment (a 4-day window of about 2.2M requests; aggregate serving-stack
histograms only, with no request content or session data) against WEKA and four traces
captured with Exgentic on the same model (HWE-Bench, CyberGym, HANDBOOK, and DeepSWE;
Table~\ref{tab:repr-main}). Each of these traces was closer to the typical agentic workload than WEKA in both input and output lengths, yet none matched it
on its own, which supports capturing several use-case benchmarks rather than
relying on a single public trace (more details in Appendix~\ref{app:repr}).

\begin{table}[t]
\caption{KS distance between per-request token-length distributions of each workload
and typical agentic workload on the GLM-5.3 deployment (lower is closer).}
\label{tab:repr-main}
\centering
\small
\begin{tabular}{lcc}
\toprule
Workload & Input tokens & Output tokens \\
\midrule
HWE-Bench  & \textbf{0.135} & \textbf{0.044} \\
CyberGym   & 0.140 & 0.133 \\
HANDBOOK   & 0.160 & 0.164 \\
DeepSWE    & 0.161 & 0.081 \\
WEKA       & 0.223 & 0.212 \\
\bottomrule
\end{tabular}
\end{table}

\subsection{Lessons Learned}
\label{sec:lessons}

Each tool closed a gap the previous step left open. Public traces and a load ladder gave a starting configuration but described a different workload; Exgentic captured the intended use cases with full content, the trace format made them replayable, and dependency-aware replay with session duplication turned a limited set of recorded sessions into high-concurrency load. Together they moved two problems, a capacity shortfall and a routing imbalance, from production into sizing and pre-production, where they are cheap to fix. Table~\ref{tab:lessons} summarizes the lessons. 

\begin{table*}[t]
\centering
\footnotesize
\caption{Lessons from the llm-d bring-up, with the evidence for each.}
\label{tab:lessons}
\begin{tabularx}{\textwidth}{@{}p{0.22\textwidth}X@{}}
\toprule
\textbf{Lesson} & \textbf{Evidence} \\
\midrule
\textbf{Public traces suffice for initial configuration, not for sizing.}
  & Tuning on WEKA and Moontrace gave a working configuration, but captured workloads reached ${\sim}0.5\times$ the optimal concurrency and ${\sim}60\%$ of the throughput (Sizing, step 1).
   \\
\addlinespace
\textbf{Record and replay on the target model.}
  & Replay fidelity is highest when the original model is used, since its responses are closer to the recorded version, whereas equivalent fidelity cannot be expected from traces generated by other models.
  \\
\addlinespace
\textbf{No single benchmark matches the typical agentic workload.}
  & Every captured trace was closer to the typical agentic workload than WEKA, yet none matched it alone (Table~\ref{tab:repr-main}).
   \\
\addlinespace
\textbf{Router bugs depend on the prefix structure of the workload.}
  & The routing imbalance appeared only with a constant shared prompt prefix, which WEKA lacks (Pre-production $\rightarrow$ sizing, step 2).
  \\
\bottomrule
\end{tabularx}
\end{table*}

%% file: related_work.tex
\section{Related Work}
\label{sec:related_work}

\paragraph{Agentic Platform Benchmarking.}
Serving providers describe how they evaluate their own platforms mostly in blog
posts. Baseten's day-0 Kimi K3 deployment~\citep{baseten2026kimik3} follows
stages close to ours: quality checks (Grade School Math 8K
(GSM8K)~\citep{cobbe2021gsm8k}, the Berkeley Function-Calling Leaderboard
(BFCL)~\citep{patil2025bfcl}, and the Kimi Vendor
Verifier~\citep{kimivendorverifier}, rerun as a regression check), configuration sweeps over parallelism,
batch size, speculation, and routing across vLLM, SGLang, and an in-house
engine, and finally scale-out with KV-aware routing. However, it reports no quality scores or end-to-end latency or throughput, and does not specify the workloads behind its sweeps, optimization, or scale validation, so it is unclear whether these stages use distinct representative workloads.

The llm-d project~\citep{llmd2025kvcache} runs a controlled
prefix-cache routing benchmark on 16 H100s using synthetic multi-tenant
traffic with shared prefixes rather than agentic sessions. The llm-d 1.0
announcement~\citep{google2026llmd} reports headline agentic coding results
and links to a Qwen3-Coder guide and Prism that document the workload,
deployment configuration, and concurrency sweep, but its workload is also
synthetic (Inference Perf) rather than captured agent traces, and it reports
no run-to-run variance.
For quality, Moonshot~\citep{kimivendorverifier} and
MiniMax~\citep{minimaxproviderverifier} publish verifiers that third-party
hosts run against their deployments, but these are model-specific. None of
these efforts drives sizing with replayed agentic traces or links the stages
of the release cycle; our functional API tests are model-agnostic and cover the
full deployed path, including the gateway.

\paragraph{Trace Replay.}
Because agent behavior depends on nondeterministic LLM outputs, identical tasks can generate different requests, making serving-system comparisons unfair. XPerf~\citep{wang2026xperf}
addresses this with fine-grained replay of traces captured from real agentic
applications, reproduced across serving systems. It presents output token
ID enforcement, which is prototyped in vLLM.
AgentReplay~\citep{pan2026agentreplay} further argues that greedy decoding and
length-only replay are insufficient, since token content drives prefix-cache
hits and mixture-of-experts routing. It therefore forces the recorded output
tokens, and optionally expert selections and tool delays, during otherwise
normal decoding. Token-wise forcing requires control inside each engine's
decoding loop and router, which is impractical for Inference
Perf: it targets many inference engines
(vLLM, SGLang, and others) through their standard serving APIs. We instead
make replay output-aware: live outputs are propagated to dependent requests,
and recorded tool choices are enforced through the API's tool-choice
parameter, which keeps tool-call behavior aligned with the trace on any
compatible engine (Section~\ref{sec:trace-replay}).

Our work is complementary: it connects quality evaluation,
workload capture, scalable replay, controlled stress testing, and regression
analysis across the deployment lifecycle.

%% file: conclusion.tex
\section{Conclusion}
\label{sec:conclusion}

Bringing an agentic serving platform from a first deployment to production requires more than running a single benchmark. Instead, deployment proceeds through a sequence of stages, each asking a different
question and needing different data. We presented an end-to-end evaluation
pipeline that follows this lifecycle and connects its stages through a single
replayable trace format. Live benchmarking with real agents records every
model call in full, so the runs that measure quality also produce the traces
that drive sizing; the same format can also carry production traces back into the pipeline upon user consent.

Dependency-aware replay preserves the per-session load characteristics of recorded agent sessions: the 95th-percentile input differs by at most $1.18\times$ from the recording, versus up to $4.98\times$ under live re-runs, and vLLM statement coverage
shows that the evaluated live, replayed, and synthetic workloads reach
complementary Python serving paths in this configuration. In the bring-up of four large MoE models on llm-d, traces captured on the target model reached about half the optimal concurrency, and correspondingly about 60\% of the throughput, of public traces, and exposed a routing imbalance before production. Experiments on the GLM-5.3 deployment 
showed that the captured workloads are
closer to typical agentic traffic than public traces.

Large-scale evaluation of synthetic workload generation remains future work. We contributed the replay and synthetic workload
tools in Inference Perf, the Exgentic extensions for live benchmarking, and a
dataset of 10{,}056 agent sessions with full request content.

%% file: ai_disclosure.tex
\section{AI Usage Disclosure}
\label{sec:ai_disclosure}

AI is both the subject of this work and a tool used throughout it. The platform we evaluate serves large language models, and our experiments exercise GLM-5.3, Nemotron~3~Ultra, Kimi~K3, and DeepSeek~V4.1~Flash (Sections~\ref{sec:live-benchmark}--\ref{sec:usecase}). The rest of this section describes AI as a tool.

\textbf{How we worked with AI.} Each author worked with AI coding assistants throughout the project, mainly Claude Code and Codex. An assistant typically supported the same author across every phase of the work: surveying existing tools, trace formats, and literature; designing experiments and metrics; implementing the tools; running experiments and analyzing their results; and drafting and reviewing text, tables, and figures, for this paper and for related artifacts such as blog posts. Because the same assistant, often across several sessions, had already worked on a tool or survey, it carried that context into later steps. For example, the comparisons of trace formats and replay mechanisms (Tables~\ref{tab:trace_formats} and~\ref{tab:replay_capabilities}) grew out of AI-assisted surveys carried out while building the tools; the tables were drafted with AI, re-verified against documentation and source code by assistant sessions that had done that survey work, and then checked by the authors. Authors also had assistants consult one another's sessions for details. Design decisions, metric definitions, and the final text were made or approved by the authors.

The remaining paragraphs describe the specific uses where AI output affects data, results, or claims, and how each is checked.

\textbf{Data generated by AI.} Exgentic itself is conventional software, but the traces it records are executions of AI agents: the released Exgentic traces (Section~\ref{sec:trace_format}) were produced by five agent harnesses, including Claude Code, driving five models to solve benchmark tasks. Two of the benchmarks involve an LLM beyond the agent under test: $\tau^2$-bench uses an LLM-based user simulator, and BrowseComp-Plus uses an LLM judge that determines the task score. These calls are removed from the released traces, but they shape the agents' trajectories and, for BrowseComp-Plus, the reported scores. The traces we captured for this paper (HWE-Bench, CyberGym, Terminal-Bench~2.1, SWE-Bench Multimodal, DeepSWE, and HANDBOOK) come from benchmarks scored by deterministic programmatic checks (unit tests or rubric verifiers), without an LLM judge or user simulator. Synthetic workloads (Section~\ref{sec:synthetic}) are generated procedurally, without a model.

\textbf{AI inside our tools at runtime.} Trace replay (Section~\ref{sec:trace-replay}) is driven by recorded traces, not by an AI agent; however, the responses of the model under test are fed into the downstream requests, so model output becomes part of the replayed workload by design. The dashboard's automated run analysis (Section~\ref{sec:dashboard}) uses Claude Code (with Claude Sonnet by default; the model is configurable) to write the interpretive part of each report (verdict, recurring errors, trajectory observations). All factual figures are computed deterministically beforehand, and a validator rejects any report whose facts differ from the precomputed values, so the model can interpret the numbers but not change them.

\textbf{AI in reported results.} The replay-fidelity and consistency metrics (Section~\ref{sec:trace-replay}) are computed by a script generated with Claude Code; the metric definitions were chosen by the authors.

\textbf{References.} AI assistants were used to find candidate references. Every reference was checked by the authors to exist and to support the claim it is cited for.

\textbf{Responsibility.} The authors take full responsibility for the content of this paper, including all text, code, data, and results produced with AI assistance.

%% file: appendix_dashboard.tex
\section{Dashboard Details}
\label{app:dashboard}

\begin{figure*}[t]
  \centering
  \includegraphics[width=\textwidth]{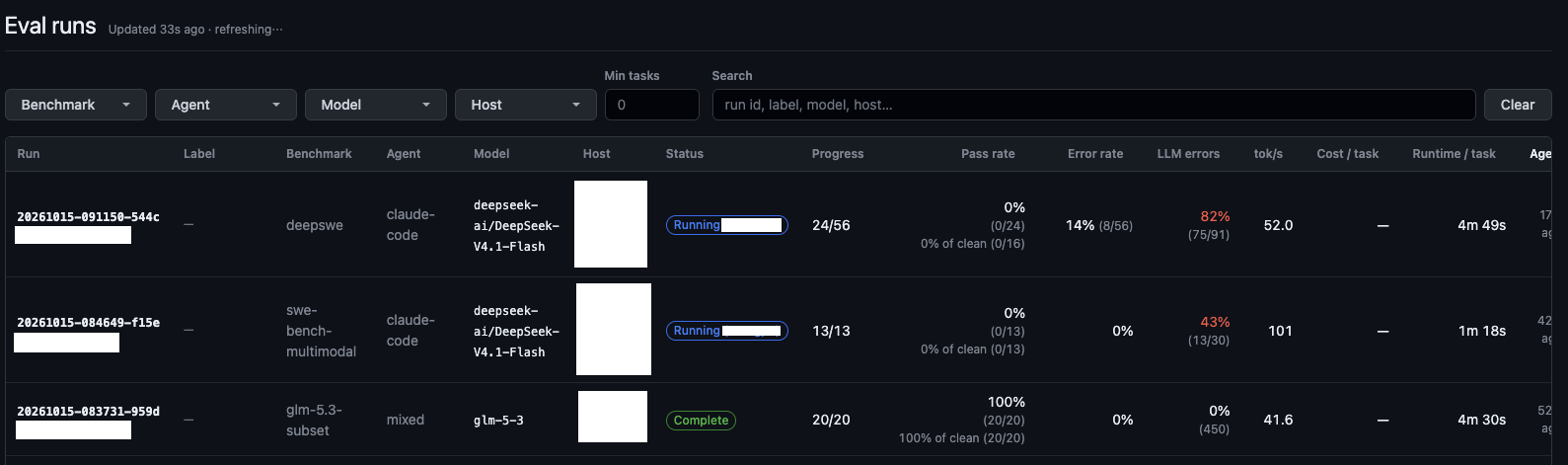}
  \caption{Monitoring view: every evaluation job on the cluster, with live progress, pass and error rates, LLM errors, throughput, cost, and runtime per task.}
  \label{fig:dashboard_runs}
\end{figure*}

\begin{figure*}[t]
  \centering
  \includegraphics[width=0.85\textwidth]{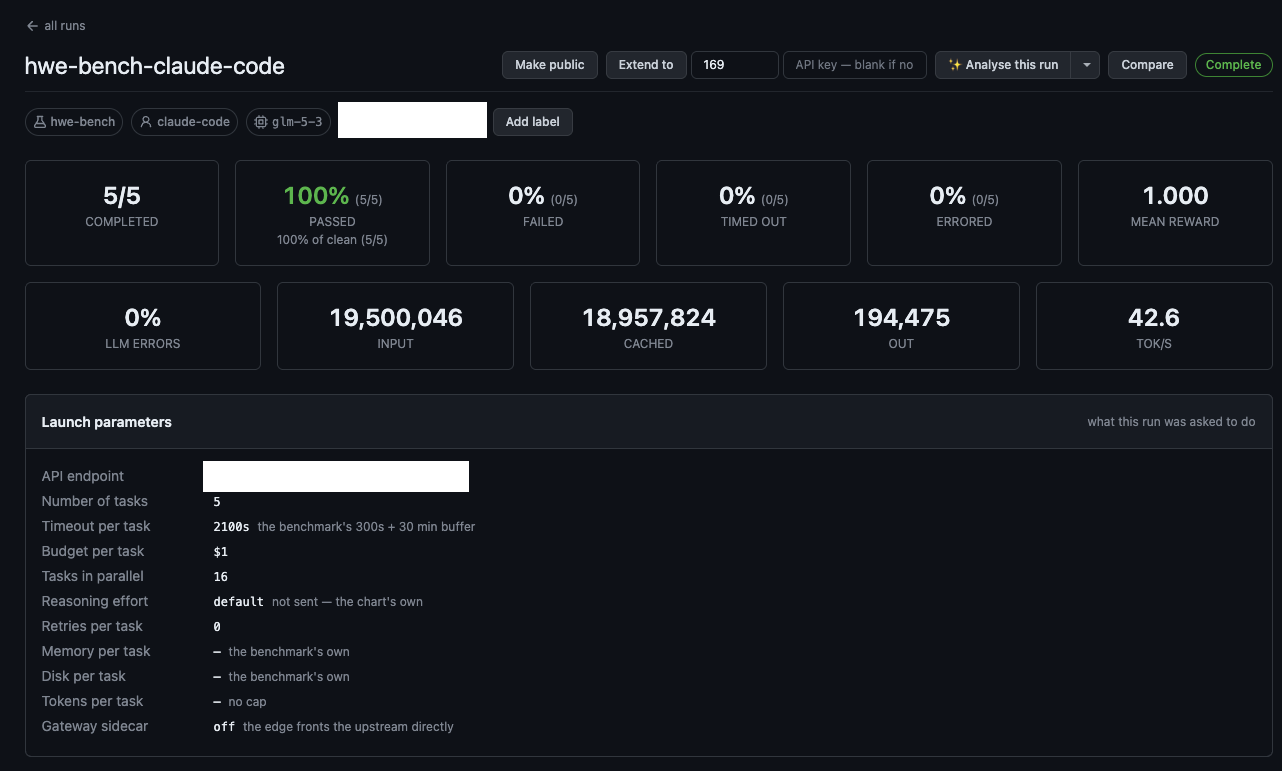}
  \caption{Run view: outcome and token-usage summary, launch parameters, and actions to publish, extend, analyze, or compare the run.}
  \label{fig:dashboard_run_detail}
\end{figure*}

\begin{figure*}[t]
  \centering
  \includegraphics[width=0.85\textwidth]{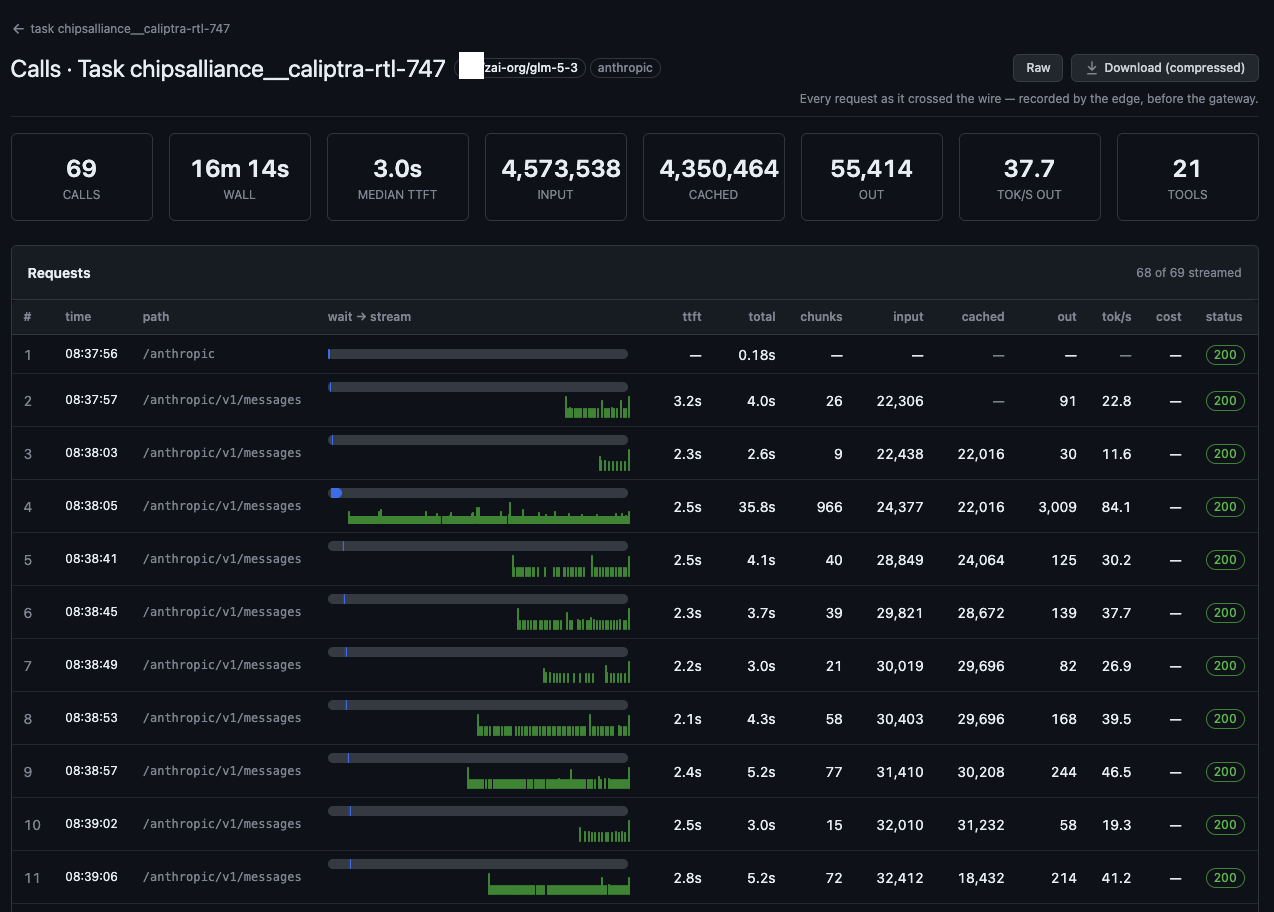}
  \caption{Task call view: every model request of a single task as recorded at the inference edge, with time to first token, latency, streaming profile, and token counts. The recorded calls can be downloaded and used for trace replay (Section~\ref{sec:trace-replay}).}
  \label{fig:dashboard_task_calls}
\end{figure*}

\begin{figure*}[t]
  \centering
  \includegraphics[width=0.85\textwidth]{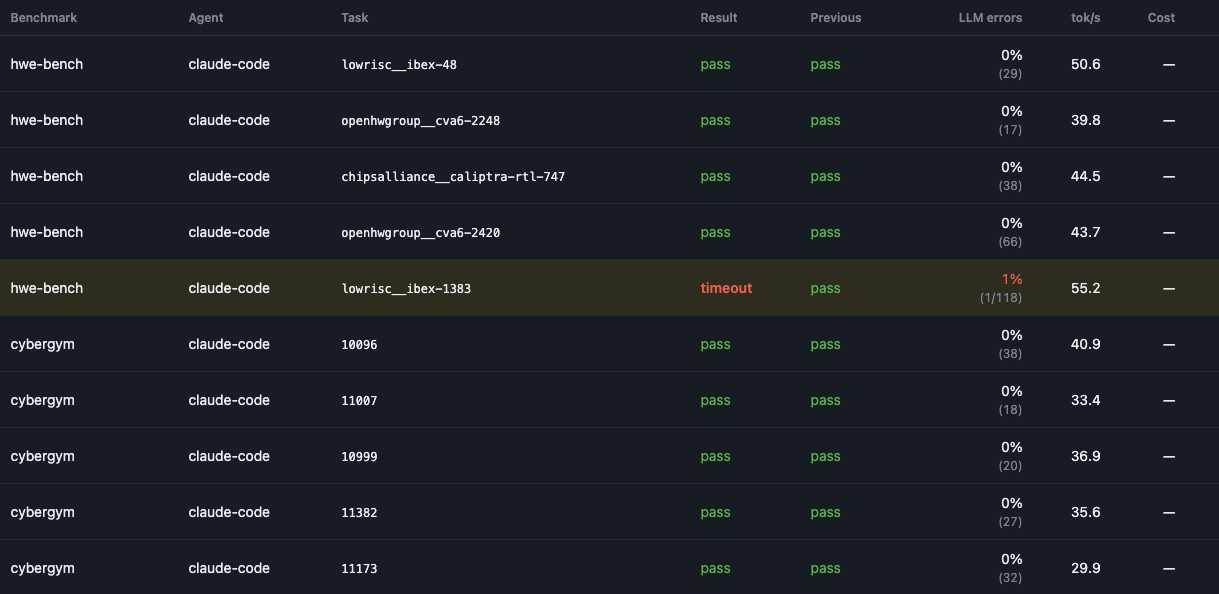}
  \caption{Regression view: A model deployment is evaluated against a fixed suite of tasks, and the results are compared with those from a previous run.}
  \label{fig:dashboard_regression}
\end{figure*}

\begin{figure*}[t]
  \centering
  \includegraphics[width=0.85\textwidth]{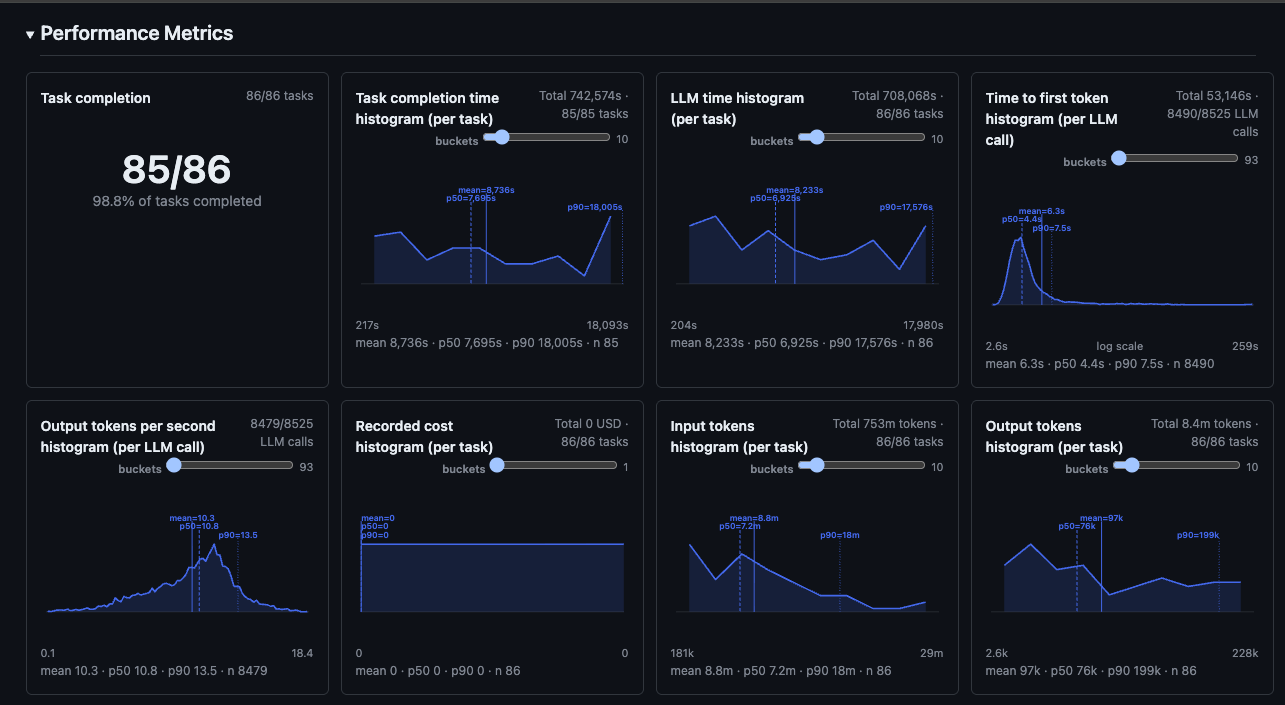}
  \caption{Analysis report, performance metrics: per-task distributions of completion time, LLM time, cost, and input and output tokens, and per-call distributions of time to first token and output throughput, each with mean, p50, and p90.}
  \label{fig:dashboard_report_metrics}
\end{figure*}

\begin{figure*}[t]
  \centering
  \includegraphics[width=0.85\textwidth]{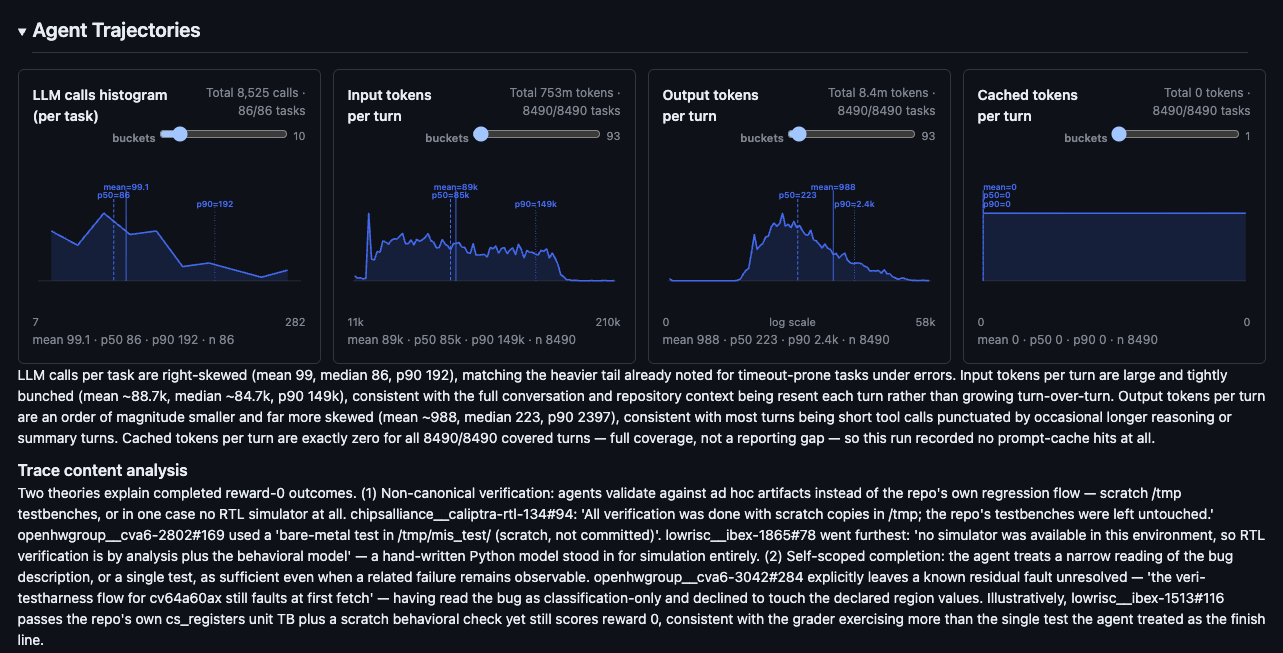}
  \caption{Analysis report, agent trajectories: per-task distribution of LLM calls and per-turn distributions of input, output, and cached tokens, followed by the agent-written interpretation of these distributions and a trace content analysis that cites specific tasks and turns.}
  \label{fig:dashboard_report_trajectories}
\end{figure*}

\paragraph{Launching.}
The launch form offers only benchmark$\times$agent combinations whose container images are published in the registry, so a user cannot select a pair that would fail to start. A UI launch renders the same Helm chart as the command-line workflow, so dashboard launches are interchangeable with CLI runs. Runs can be restricted to task subsets, extended, retried, or cancelled.

\paragraph{Monitoring and inspection.}
Each benchmark run appears in the monitoring view as a single row with live progress, pass rate, mean reward, token usage, and cost, alongside the launcher and current cluster load (Figure~\ref{fig:dashboard_runs}). The run view shows per-task reward and pass/fail status, and each task's trace records every model call with its prompt, response, tool calls, tokens, and latency. Aggregate usage is broken down into input, output, and cached tokens, latency, time to first token, throughput, and cost (Figures~\ref{fig:dashboard_run_detail}--\ref{fig:dashboard_task_calls}).

\paragraph{Regression suites.}
A suite is a fixed collection of tasks drawn from selected benchmark$\times$agent combinations. For each model, suites are seeded automatically from historical results with small, diverse sets of fast-running tasks that previously passed, giving a low-cost regression check. A suite is immutable, so two of its executions, for example across deployment versions, can be compared directly. When two executions of a suite are compared, each task is categorized as pass, fail, or error in each run, and users can drill down into any disagreement category (Figure~\ref{fig:dashboard_regression}). Running a suite against a model launches each cell as an ordinary run and records which runs belong to that suite execution.

\paragraph{Automated analysis.}
On request, a deterministic pass computes completion, latency, token and cost distributions, tool usage, and structured errors; Claude Code then adds interpretive content, such as an overall verdict, recurring errors and anomalies, and trajectory-level observations (Figures~\ref{fig:dashboard_report_metrics} and~\ref{fig:dashboard_report_trajectories}). A validator rejects any report whose facts differ from the precomputed values, so the model can interpret the numbers without altering them. A rule-based mode produces the same report without a model. Each analysis runs under a cost budget, and its own model calls, tokens, and cost are recorded and shown alongside the report. Because the report summarizes the distributions and metrics of a run's traces, it can also serve as a summary of the workload capture stage before those traces are replayed.

\paragraph{Sharing.}
The dashboard has a public and an internal view. Runs are internal by default and can be published individually; the public view lists only published runs and, for launching, offers only benchmark$\times$agent pairs that have a published run. Reports and comparisons can be exported as standalone HTML or JSON.

\paragraph{Implementation.}
The dashboard is a FastAPI/htmx web application deployed on OpenShift. It keeps no database: the cluster's Jobs and the results stored in object storage (IBM COS) serve as the state, with a small per-run metadata file for the launcher, visibility, and label. Analyses and comparisons run as separate, sandboxed Kubernetes Jobs that write their artifacts next to the run's results.

%% file: appendix_replay.tex
\section{Trace Replay Details}
\label{app:replay}

Table~\ref{tab:trace_formats} compares the information retained by the trace formats discussed in Section~\ref{sec:trace_format}.

\paragraph{Trace formats.} WEKA records subagent patterns, timing, token counts, and prefix structure, but hashes message content, so requests cannot be reproduced verbatim. SWE-smith~\cite{yang2025swesmithscalingdatasoftware} captures software-engineering actions and outcomes, while TraceLab~\cite{zhu2026tracelabcharacterizingcodingagent} normalizes sanitized traces from Claude Code and Codex, omitting original user messages and tool input/output text. Mooncake~\cite{qin2024mooncake,qin2025mooncake} and Baseten~\cite{aiperf_baseten_trace} prioritize workload shape and cache behavior; Baseten retains a flattened completion prompt, but not structured messages, tool calls, or tool definitions.

\begin{figure*}[t]
  \centering
  \includegraphics[width=0.85\textwidth]{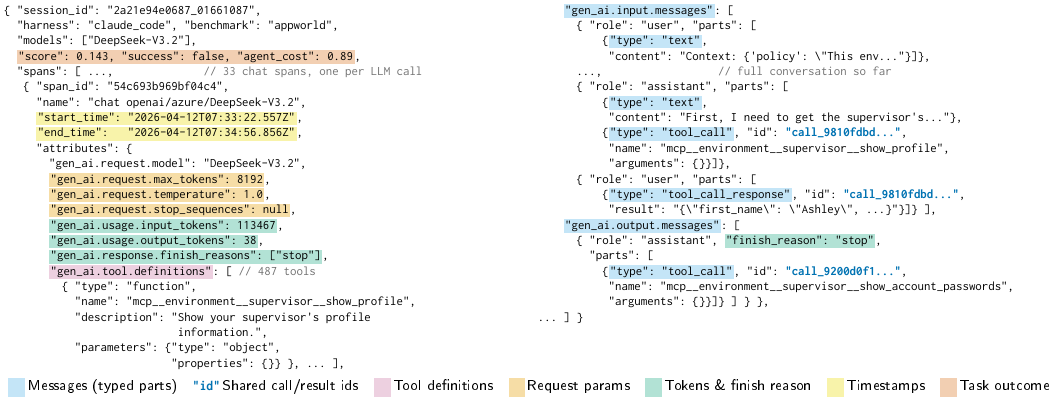}
  \caption{An example agent session from the released Exgentic trace dataset (Section~\ref{sec:trace_format}).}
  \label{fig:exgentic_example}
\end{figure*}

\paragraph{Exgentic traces.} Each Exgentic session contains one OTel GenAI chat span per LLM call, retaining complete input and output messages as typed text, tool-call, and tool-result parts; tool definitions; request parameters; token counts; finish reasons; and start and end timestamps. Each session also records its task outcome. Shared identifiers link tool calls to their results, including parallel calls. Dependencies between LLM requests are not explicit, but can be inferred from message content and timing; agent hierarchy is likewise not annotated, so subagent delegation may appear only as a tool call. The released traces retain calls from the agent under test, excluding benchmark scaffolding and failed retries, and provide call-level start and end timestamps. Figure~\ref{fig:exgentic_example} shows an example session.

\begin{table}[t]
\centering
\scriptsize
\setlength{\tabcolsep}{4pt}
\caption{Information available from each trace format (\yes: explicitly recorded; \pt: incomplete or heuristically inferable; \no: unavailable or not applicable). Msgs.: structured messages (roles and content); Calls: tool invocations; Deps.: execution dependencies between requests; Hier.: hierarchical agent relationships; Defs.: tool definitions; Params.: request parameters (e.g., max tokens, temperature); Tokens: token counts; Timing: timestamps or latencies; Prefix: a cross-request prefix-sharing or cache signal recorded in the trace itself (e.g., per-block prefix hashes or cached-token counts), not merely content from which one could be computed.}
\label{tab:trace_formats}
\begin{tabular}{lccccccccc}
\toprule
Format & Msgs. & Calls & Deps. & Hier. & Defs. & Params. & Tokens & Timing & Prefix \\
\midrule
AgentX (WEKA) & \no & \no & \pt & \yes & \no & \no & \yes & \yes & \yes \\
SWE-smith & \yes & \yes$^{a}$ & \no & \no & \no & \no & \no & \no & \no \\
TraceLab & \no & \pt & \pt & \pt & \no & \no & \yes & \yes & \yes \\
Mooncake & \no & \no & \no & \no & \no & \no & \yes & \yes & \yes \\
Baseten & \pt & \no$^{b}$ & \pt & \no & \no$^{b}$ & \no & \yes & \yes & \yes \\
\midrule
\textbf{Exgentic} & \yes & \yes$^{c}$ & \pt & \no & \yes & \yes & \yes & \yes & \no \\
\bottomrule
\end{tabular}

\vspace{2pt}
\parbox{\columnwidth}{\scriptsize
$^{a}$Only in the \texttt{tool} split; the \texttt{xml} and \texttt{ticks} splits flatten tool calls into message text.
$^{b}$Not applicable: Baseten traces replay \texttt{/v1/completions}, which has no tool-calling interface.
$^{c}$Recorded as typed tool-call and tool-result message parts with shared identifiers; tool executions have no spans of their own, so tool latency is not recorded.}
\end{table}

Table~\ref{tab:replay_capabilities} compares AIPerf,\footnote{\url{https://github.com/ai-dynamo/aiperf/blob/main/docs/benchmark-modes/trace-replay.md}} Inference Perf,\footnote{\url{https://github.com/kubernetes-sigs/inference-perf/blob/main/docs/otel_trace_replay.md}} GuideLLM,\footnote{\url{https://github.com/vllm-project/guidellm/blob/main/docs/en/guides/trace_replay.md}} vLLM bench serve,\footnote{\url{https://github.com/vllm-project/vllm/blob/main/docs/benchmarking/cli.md}} SGLang \texttt{bench\_serving},\footnote{\url{https://github.com/sgl-project/sglang/blob/main/docs/docs/developer_guide/bench_serving.mdx}} and NVIDIA DynoSim.\footnote{\url{https://github.com/ai-dynamo/dynamo/blob/main/docs/fern/pages/reference/components/dynosim-replay-cli-reference.mdx}}

\input{replay_capabilities_table}

\paragraph{Format and loader behavior.} Basic Mooncake rows preserve timestamps, token counts, and prefix hashes, enabling serving-shape replay with synthesized prompts but not workflow reconstruction. AgentX (WEKA) traces add declared subagent structure; AIPerf, GuideLLM, and DynoSim use it to reconstruct spawn/join topology, with some adapters also inferring flattened branches from prefix hashes and timing. These formats may therefore preserve graph shape without original prompt and response bodies. Exgentic traces instead retain OTel GenAI spans with full request snapshots, recorded responses, tool definitions, request parameters, and timestamps. Inference Perf infers dependencies using message/output matches and temporal fallback, propagates live responses, and preserves recorded shared-prefix, tool-call, and image structure. Its loader accepts span JSON from files, directories, or Hugging Face datasets, and sessions can be duplicated with cache-isolating markers to scale load without artificial prefix sharing.

\paragraph{Replay-system distinctions.} Capability labels in Table~\ref{tab:replay_capabilities} do not imply equivalent implementations. AIPerf and GuideLLM provide graph replay for WEKA traces, while their other loaders have different semantics. SGLang propagates live responses in sequential \texttt{agentic-trace} mode, whereas timestamped Mooncake replay uses placeholder assistant responses. DynoSim preserves agentic dependency graphs but executes them against simulated rather than live engines. Inference Perf combines OTel ingestion, recorded timing, dependency-aware parallel and fan-in scheduling, live-output propagation, and a live inference target in one path.

\begin{figure*}[t]
  \centering
  \includegraphics[width=\textwidth]{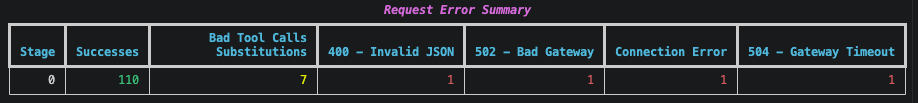}
  \caption{Per-load-condition error report emitted at the end of a trace replay run (Section~\ref{sec:trace-replay}): successful requests, bad tool-call substitutions in which the recorded assistant message replaced a live response with malformed tool-call arguments, and failed requests grouped by cause.}
  \label{fig:replay_error_summary}
\end{figure*}

\paragraph{Error reporting.} When responses are propagated, a request failure affects every downstream request. We contribute to Inference Perf a report for each configured load condition (``stage''), emitted at the end of the run (Figure~\ref{fig:replay_error_summary}), that lists successful requests and failures grouped by HTTP status or transport error.\footnote{\url{https://github.com/kubernetes-sigs/inference-perf/blob/main/docs/reports.md}} It also counts \emph{bad tool-call substitutions}, where malformed live tool-call arguments cause the recorded assistant message to be substituted, and reports failed sessions and downstream events cancelled because a predecessor failed.

\subsection{Replay Evaluation Details}
\label{app:replay_eval}

\paragraph{Workflow fidelity.}
\[ \mathrm{WF} = 1 - \frac{\mathrm{Lev}(S_{\mathrm{rec}}, S_{\mathrm{rep}})}{\max(|S_{\mathrm{rec}}|, |S_{\mathrm{rep}}|)}, \]
where $S$ is the session call-type sequence $[\texttt{user}, \texttt{assistant}, \texttt{tool}, \ldots]$. Each LLM invocation contributes one \texttt{assistant} symbol, followed by one \texttt{tool} symbol for every tool call it generates. A \texttt{user} symbol is emitted when a conversation opens, or after an LLM call that invokes no tools.

\paragraph{Tool-call fidelity.}
\[ \mathrm{TF} = \frac{|T_{\mathrm{rec}} \cap T_{\mathrm{rep}}|}{\max(|T_{\mathrm{rec}}|, |T_{\mathrm{rep}}|)}, \]
computed over the multisets of invoked tool names and averaged over the turns in which the recorded or the replayed trace issues at least one tool call. WF and TF are averaged across replayed sessions.

\paragraph{Load ratios.} For each session $s$ with turns $i$, we compare the replayed (or re-run) load to the recorded load:
\[ R_{\mathrm{in},s} = \frac{\sum_i \mathrm{in}^{\mathrm{rep}}_{s,i}}{\sum_i \mathrm{in}^{\mathrm{rec}}_{s,i}}, \]
\[ R_{\mathrm{out},s} = \frac{\sum_i \mathrm{out}^{\mathrm{rep}}_{s,i}}{\sum_i \mathrm{out}^{\mathrm{rec}}_{s,i}}, \]
\[ R_{\mathrm{peak},s} = \frac{\max_i \mathrm{in}^{\mathrm{rep}}_{s,i}}{\max_i \mathrm{in}^{\mathrm{rec}}_{s,i}}. \]
Input tokens are the server-reported prompt size, including cached tokens. Output tokens are the server-reported completion count, including reasoning. Each side is aggregated over all of its own turns rather than over paired turns, so turns that a replay drops or adds count toward its load. A ratio of $1\times$ means the session placed the same load on the server as the recording. We report the median and the 95th percentile (P95, linearly interpolated) over the 40 sessions of each benchmark (Table~\ref{tab:replay_load_ratio}).

At the median, replay and live re-runs both stay near $1\times$. The one exception is replay output on HWE-Bench ($0.74\times$): replayed GLM-5.3 turns generate fewer tokens than the recorded ones, while replayed input and peak context stay at $0.91\times$ and $0.89\times$. The two modes diverge in the tail. A re-run session can take a different path, e.g., more steps or different tool outputs. These differences compound to up to $4.98\times$ the recorded input, $4.52\times$ the recorded output and $1.85\times$ the recorded peak context. Replay reuses the recorded tool results and constrains tool selection (Section~\ref{sec:trace-replay}), so its P95 input and peak-context ratios stay within $1.12$--$1.30\times$. Output is the least constrained quantity under replay, because only the structure of each turn is enforced, not its length. Even so, its P95 ratio of $1.28$--$2.04\times$ is less than half of the re-run tail.

\paragraph{TPOT CV.} $\mathrm{CV} = \sigma/\mu$, the sample standard deviation over the mean of per-replay TPOT. TPOT does not depend on how much each replay generates, so it should be stable on a stable server even though live output substitution changes the outputs.

\paragraph{Workflow consistency.}
\[ \mathrm{WC} = \frac{2}{R(R-1)} \sum_{i<j} \left( 1 - \frac{\mathrm{Lev}(S_i, S_j)}{\max(|S_i|, |S_j|)} \right), \]
the mean over all pairs of the $R$ replays, where $S_i$ is replay $i$'s call-type sequence, built as for WF over its successful LLM calls. $\mathrm{WC}_{\mathrm{role}}$ uses the call-type symbols only. $\mathrm{WC}_{\mathrm{tool}}$ also labels each \texttt{assistant} symbol with the names of the tools it called. A score of 100\% means every replay followed the same workflow.

\paragraph{Consistency setup.} The two traces come from GLM-5.3 on HWE-Bench task \emph{chipsalliance\_\_caliptra-rtl-594} and DeepSeek-V4.1-Flash-1017 on CyberGym task \emph{1237}. Each is replayed 50 times with Inference Perf, giving 1{,}225 replay pairs per trace.

%% file: replay_capabilities_table.tex
\begin{table*}[t]
\centering
\scriptsize
\caption{Capabilities of representative trace-replay mechanisms, based on documentation and source code accessed October 21, 2026 (\yes: supported; \pt: limited or approximated; \no: unsupported). Cells report tool-level capability: capabilities in the same column may be provided by different loaders or modes and need not be composable in one replay path. ``Timing'' means scheduling from recorded timestamps or inter-call gaps; ``workflow'' means replay beyond independent request rows. DynoSim targets simulated engine cores, whereas the other mechanisms drive live endpoints.}
\label{tab:replay_capabilities}
\setlength{\tabcolsep}{4pt}
\begin{tabular}{lcccccc}
\toprule
\textbf{Capability} & \textbf{AIPerf} & \textbf{Inference Perf} & \textbf{GuideLLM} & \textbf{vLLM} & \textbf{SGLang} & \textbf{DynoSim} \\
\midrule
Recorded timing                 & \yes & \yes & \yes & \yes & \yes & \yes \\
Conversation replay             & \yes & \yes & \yes & \yes & \yes & \yes \\
Live output propagation         & \yes & \yes & \yes & \yes & \yes & \no \\
Dependency-aware scheduling     & \yes & \yes & \yes & \no & \no & \yes \\
Parallel branches               & \yes & \yes & \yes & \no & \no & \yes \\
Fan-in synchronization          & \yes & \yes & \yes & \no & \no & \yes \\
Workflow-level replay           & \yes & \yes & \yes & \pt & \pt & \yes \\
Native OTel input               & \yes & \yes & \yes & \no & \no & \no \\
Live inference target           & \yes & \yes & \yes & \yes & \yes & \no \\
\bottomrule
\end{tabular}
\vspace{2pt}

\parbox{\textwidth}{\footnotesize\emph{Context.} When this work began, no other surveyed system publicly supported OTel replay. AIPerf added Weka/DAG and Exgentic replay roughly one and three months later; GuideLLM added comparable capabilities roughly four to six months later. This chronology establishes public availability, not design provenance, and the workflow is replay-system agnostic.}
\end{table*}

%% file: appendix_synthetic.tex
\section{Synthetic Workload Generation Details}
\label{app:synthetic}

This appendix details the synthetic agentic workload generator summarized in Section~\ref{sec:synthetic}.

\paragraph{Controllable trace patterns.} Sessions may consist of multiple conversational turns, repeated tool-calling loops, parallel tool calls, and recursive sub-agent spawning. Context evolves naturally as conversations progress, causing prompt lengths and KV-cache utilization to grow over time. The generator also supports explicit context compaction, where accumulated history is replaced by a shorter summary once a configurable threshold is reached, enabling evaluation of systems under both context growth and context-reset scenarios.

\paragraph{Themes.} Prompt sizes, response lengths, tool catalog sizes, and interaction latencies are controlled through statistical distributions, while a \textit{theme} supplies the content. A theme is a declarative description of a domain (e.g., software engineering, retrieval-augmented generation, or incident response) consisting of a system prompt, a tool catalog with descriptions and parameter schemas, per-tool result templates, an objective template, introductory documents (e.g., an issue report), and domain filler snippets (e.g., code, log, and CI lines). Each session draws a theme from a weighted mix and pins one value per entity category (service, symptom, file, symbol, test), so every turn refers to the same subject: the first user turn presents the objective and an introductory document, tool results are rendered from the tool's template, and filler pads each turn to its sampled token budget. Token counts therefore follow the configured distributions, while the prompts stay semantically coherent, so the model under test emits plausible tool calls and answers rather than reacting to random text.

\paragraph{Graph construction.} The generator emits a \textit{ReplayGraph}, the same DAG representation used by trace replay, so synthetic workloads exercise the identical scheduling, dependency, output-substitution, and metric-collection paths, keeping comparisons with trace-driven runs meaningful. Each node represents a single LLM invocation. During a tool loop, the model produces one or more tool calls, receives synthetic tool outputs, and iterates until a final answer is emitted. Fan-out patterns are modeled through \texttt{dispatch\_agent}-style actions that create parallel child agents, each executing its own workflow before returning results to a parent aggregation step. All stochastic decisions, including branching, tool-loop depth, token counts, and sub-agent creation, are derived from a configuration-controlled random seed, making generated workloads fully deterministic and reproducible. Figure~\ref{fig:synthetic-example} shows an abbreviated synthesized session.

\begin{figure}[t]
\centering
\begin{lstlisting}[style=trace]
SYSTEM: You are an autonomous software-engineering agent ... tools=[list_dir, read_file, grep_code, find_symbol, git_diff, run_tests, run_command, edit_file]

USER: <context> def OrderHandler(cart, ...) ... pytest tests/test_cache_... FAILED </context> Issue #474 (orders-api) Title: incorrect tax rounding File "src/billing/tax.py", function apply_discount AttributeError: 'NoneType' has no attribute 'amount'

ASSISTANT: grep_code("apply_discount")

TOOL: 78 matches in orders-api src/billing/tax.py:516: total = total - ...

ASSISTANT: find_symbol("apply_discount")

TOOL: apply_discount -> src/billing/tax.py:105

ASSISTANT: ...

TOOL: ...

USER: Final turn. Answer now.

ASSISTANT: <final answer, 80 tokens> \end{lstlisting}
\caption{Abbreviated synthetic session (theme \texttt{code\_change\_task}), configured with \texttt{tool\_loop\_depth=3} which generates 3 tool calls and tool results. Pinned entities (\texttt{orders-api}, \texttt{apply\_discount}, \texttt{src/billing/tax.py}) recur throughout the interaction.}
\label{fig:synthetic-example}
\end{figure}

%% file: appendix_coverage.tex
\clearpage
\section{Workload Complementarity: Coverage Tables}
\label{app:coverage}

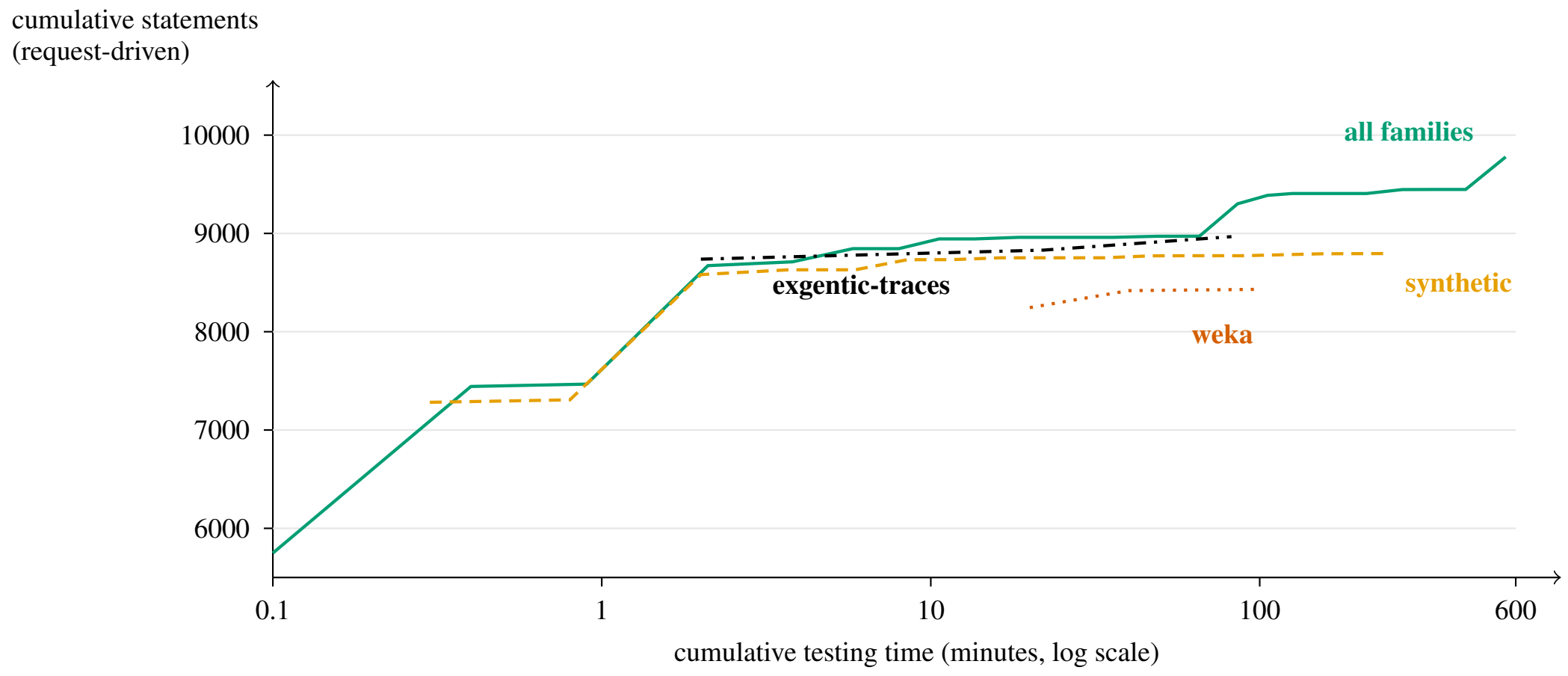
\begin{figure*}[tp]
\centering
\begin{tikzpicture}
\begin{scope}
\draw[->] (0,0) -- (11.4,0);
\node[font=\scriptsize,anchor=north] at (5.7,-0.45) {cumulative testing time (minutes, log scale)};
\draw[->] (0,0) -- (0,4.4) node[above left,font=\scriptsize,align=left] {cumulative statements\\(request-driven)};
\foreach \x/\lab in {0.000/0.1, 2.911/1, 5.823/10, 8.734/100, 11.000/600}
  \draw (\x,0) -- (\x,-0.08) node[below,font=\scriptsize] {\lab};
\foreach \y/\lab in {0.4348/6000, 1.3043/7000, 2.1739/8000, 3.0435/9000, 3.9130/10000}
  \draw (0,\y) -- (-0.08,\y) node[left,font=\scriptsize] {\lab};
\foreach \y in {0.4348,1.3043,2.1739,3.0435,3.9130}
  \draw[gray!20] (0,\y) -- (11.0,\y);

\draw[thick,markyes] plot coordinates {(0.000,0.2157) (1.753,1.6896) (2.778,1.7104) (3.850,2.7583) (4.600,2.7922) (5.134,2.9087) (5.541,2.9087) (5.897,2.9948) (6.212,2.9948) (6.601,3.0087) (6.898,3.0087) (7.152,3.0087) (7.436,3.0087) (7.830,3.0183) (8.201,3.0191) (8.538,3.3052) (8.803,3.3800) (9.023,3.3965) (9.258,3.3965) (9.680,3.3965) (9.996,3.4313) (10.272,3.4322) (10.557,3.4322) (10.911,3.7200)};
\node[markyes,font=\scriptsize\bfseries,anchor=east] at (10.75,3.95) {all families};

\draw[thick,markpt,densely dashed] plot coordinates {(1.389,1.5496) (2.629,1.5713) (3.788,2.6791) (4.566,2.7217) (5.156,2.7217) (5.617,2.8113) (6.000,2.8113) (6.448,2.8278) (6.779,2.8278) (7.055,2.8278) (7.359,2.8278) (7.774,2.8452) (8.160,2.8461) (8.590,2.8461) (9.291,2.8635) (9.844,2.8652)};
\node[markpt,font=\scriptsize\bfseries,anchor=north west] at (9.90,2.82) {synthetic};

\draw[thick,markno,dotted] plot coordinates {(6.699,2.3870) (7.576,2.5374) (8.734,2.5487)};
\node[markno,font=\scriptsize\bfseries,anchor=north east] at (8.80,2.36) {weka};

\draw[thick,black,dashdotted] plot coordinates {(3.788,2.8157) (6.820,2.8957) (8.484,3.0148)};
\node[black,font=\scriptsize\bfseries,anchor=north west] at (4.30,2.80) {exgentic-traces};
\end{scope}
\end{tikzpicture}
\caption{Cumulative request-driven statement coverage against cumulative
testing time. Within each series the runs are ordered shortest-first, so the
$x$ value of a point is the time that series had spent once that run finished;
because per-run coverage windows are independent of run order
(Section~\ref{app:coverage-method}), the endpoint of each curve is the same set
whatever order is chosen. The upper curve is all 24 runs, the lower three the
multi-run families. The time axis is logarithmic: the first order of magnitude
buys most of the coverage, and the last buys almost none. \texttt{synthetic}
reaches 8{,}581 statements after 2 cumulative minutes and 8{,}795 after
240.5 minutes, so the 151 minutes of its two longest runs add
22 statements. All families together reach 9{,}406 after
126 minutes and 9{,}778 after 559; of the 372
statements added in that last factor of four, 331 come from the single
137-minute \texttt{live} run---a different workload rather than more
time. The later runs still measure performance, which
Table~\ref{tab:coverage-workloads} shows varies across them.}
\label{fig:coverage-growth}
\end{figure*}

This appendix holds the supporting material for the coverage experiments of
Section~\ref{sec:vllm_coverage}: the exact server command line
(Figure~\ref{fig:serve-cmdline}), the static reachability partition and the
per-run isolation mechanism (Section~\ref{app:coverage-method}), the individual runs and their workloads
(Table~\ref{tab:coverage-workloads}), coverage against the statements this
deployment can statically reach (Table~\ref{tab:cumulative-coverage}), what
each test family covered and the statements it alone reached
(Table~\ref{tab:unique_coverage}), the hard-to-reach scheduling, caching and
KV-offload states each workload drove the stack into
(Table~\ref{tab:load-states}), how coverage accumulates as runs are added
(Figure~\ref{fig:coverage-growth}), and the follow-up experiment that closed
the one coverage hole those 24 runs left open
(Section~\ref{app:coverage-kvrestore}).

\subsection{Server configuration and measurement method}
\label{app:coverage-method}

Figure~\ref{fig:serve-cmdline} gives the command line verbatim.  The
static reachability partition below is computed from the modules this command
line selects, so it is reproduced here in full.

\begin{figure*}[tp]
\centering
\begin{lstlisting}[style=trace]
coverage run --rcfile=/etc/coveragerc --data-file=/coverage/.coverage \
  -m vllm.entrypoints.openai.api_server \
  --model deepseek-ai/DeepSeek-V4.1-Flash-1017 \
  --served-model-name deepseek-ai/DeepSeek-V4.1-Flash-1017 \
  --tensor-parallel-size 8 --enable-expert-parallel \
  --speculative-config '{"method":"dspark","num_speculative_tokens":5,
                         "draft_sample_method":"probabilistic"}' \
  --tokenizer-mode deepseek_v41 --reasoning-parser deepseek_v41 \
  --tool-call-parser deepseek_v41 --enable-auto-tool-choice \
  --mm-processor-cache-type shm --allowed-media-domains deny-all.invalid \
  --enable-prompt-tokens-details --max-model-len 1048576 \
  --gpu-memory-utilization 0.90 --safetensors-load-strategy prefetch \
  --enable-ep-weight-filter \
  --compilation-config '{"max_cudagraph_capture_size":64}' \
  --engram-config '{"cpu_offload":true}' \
  --kv-transfer-config '{"kv_connector":"OffloadingConnector",
    "kv_role":"kv_both","kv_connector_extra_config":{
    "spec_name":"TieringOffloadingSpec","cpu_bytes_to_use":549755813888,
    "blocks_per_chunk":4}}' \
  --block-size 64 --host 0.0.0.0 --port 8000
\end{lstlisting}
\caption{The vLLM server command line for every run in these experiments,
wrapped in coverage.py. \texttt{dspark} is the speculative-decoding method named
in \texttt{--speculative-config}; we use that spelling throughout rather than a
display form. The reachability partition is computed from the modules this
command line selects.}
\label{fig:serve-cmdline}
\end{figure*}

\paragraph{Static reachability partition.}
\label{app:coverage-reach}
Starting from the modules the command line in Figure~\ref{fig:serve-cmdline}
selects, we walk \texttt{import} edges with \texttt{ast}, resolving imports and
following the lazy registries that map a configuration string to a module. Of
the 419{,}671 measurable statements, 180{,}352 (43.0\%) sit in files this
deployment cannot import at all: 109{,}926 belong to other model
architectures, 66{,}335 to features the command line does not enable, and
4{,}091 to the served architecture built for a different accelerator.
Table~\ref{tab:cumulative-coverage} reports coverage per class. The
unreachable classes act as a control: no run executes any statement in them.

\paragraph{Per-run isolation.} To allow measuring coverage per run, without requiring a full redeploy of the image, which can take several minutes per run, we leverage coverage.py's \emph{dynamic contexts}
with \texttt{ctrace} core. Before a run's traffic starts, every measured process is switched into
a named context; when the run ends, all processes are switched into a unique
internal idle context and the data is flushed.  The per-run figures are therefore independent of run
order. 

Two caveats follow from the mechanism. Attribution is per-process at
signal-delivery time, so a process may execute a few lines---an idle loop, a
log line, the tail of an in-flight request---between the harness ending a run's
traffic and that process handling the switch; those lines are attributed to the
run just ended. And a run's window captures lazy first imports and one-time
construction as well as traffic: \texttt{exgentic-2m-c10} was the first
measured window after engine start, so some of its statements are
initialization rather than traffic-driven.

\paragraph{Excluding engine initialization from unique coverage.} The second
caveat distorts unique coverage directly, so we correct for it. The
structured-output subsystem (\texttt{vllm/v1/structured\_output/}, together
with the \texttt{LazyLoader} construction in \texttt{vllm/utils/import\_utils.py})
is built when the engine starts, regardless of whether any request asks for a
grammar. Its lines appear in whichever run holds the first measured window, and
the code involved confirms the reading: the covered statements are module
imports, class-body field declarations, backend \texttt{\_\_init\_\_} bodies,
and a request-validation function that returns immediately when a request
carries no structured-output parameters. No family in this set sends
structured-output requests, and no grammar was compiled. We therefore exclude
these 109 statements from the unique column of
Table~\ref{tab:unique_coverage}; without the exclusion they would be credited
to \texttt{exgentic-traces} as 109 of 136 unique statements, implying
that trace replay exercises three grammar backends it never invoked. The
exclusion is applied uniformly to every family, and affects only this one; the
coverage column and all cumulative figures are left untouched.

\paragraph{The three caches, in plain terms.} Table~\ref{tab:load-states}
touches three different stores, and they are easy to confuse:

\begin{itemize}
  \item \textbf{The GPU KV cache} holds the attention keys and values for
        tokens already processed, in GPU memory, so a model does not recompute
        them for every new token. It is divided into fixed-size \emph{blocks}
        (64 tokens here). Blocks no longer referenced by any request go on a
        \emph{free queue} to be handed out again.
  \item \textbf{The prefix cache} is an index over those blocks, keyed by a
        hash of the tokens they contain. When a new request begins with the
        same tokens as an earlier one---the same system prompt, or the same
        conversation so far---the server reuses the cached blocks instead of
        recomputing them. This is why agent workloads matter: every turn
        resends the whole transcript, so almost all of it is a repeated prefix.
  \item \textbf{The CPU offload tier} is a second, larger pool in host RAM
        (512\,GiB here). When the GPU cache fills, blocks are copied out to
        CPU rather than discarded, so they can be brought back if needed. That
        tier fills too, and then it must \emph{evict}: an LRU policy decides
        which blocks to drop.
\end{itemize}

A single request exercises almost none of this. It has no earlier prefix to
reuse, never fills the GPU cache, so nothing is offloaded and nothing is
evicted. The states below are the transitions that only appear once several
sessions compete for the same finite cache.

\begin{table*}[tp]
\centering
\scriptsize
\caption{Workloads used in the coverage experiments, grouped by test family.
All targeted DeepSeek-V4.1-Flash-1017 served by vLLM on eight H200 GPUs with
tensor parallelism, expert parallelism, \texttt{dspark} speculative decoding,
and GPU-to-CPU KV offloading; Figure~\ref{fig:serve-cmdline} gives the exact
command line. Duration is measured load time from each run's own load
metadata; the \texttt{cybergym} run has none, so its duration is wall-clock and
includes model load and client setup, which is why we report no per-minute
rate. Cov.\ is the statements that run executed while its own coverage
window was open, measured in isolation. The union row counts
distinct statements, so it is smaller than the column sum. The \texttt{weka}
family replays shape-only traces, and the \texttt{synthetic} family runs
synthetic agentic workloads (Section~\ref{sec:synthetic}).}
\label{tab:coverage-workloads}
\begin{tabular}{@{}p{0.15\textwidth}p{0.085\textwidth}rr p{0.515\textwidth}@{}}
\toprule
\textbf{Run} & \textbf{Family} & \textbf{Duration (min)} & \textbf{Cov.} & \textbf{Workload} \\
\midrule
\texttt{smoke}
  & smoke
  & 0.1
  & 5{,}748
  & Single non-streaming chat request (17 prompt tokens, 128 max tokens, temperature 0). Cold-start and single-call baseline. \\
\addlinespace
\texttt{cybergym}
  & live
  & 136.8
  & 8{,}777
  & Live CyberGym agent benchmark on the Anthropic messages API---the only workload not using the OpenAI chat API. Large enough KV footprint to fill the CPU offload tier and force eviction. \\
\addlinespace
\texttt{weka-\allowbreak 20m-\allowbreak c8}
  & weka
  & 20.0
  & 8{,}245
  & AIPerf streaming chat replay of WEKA multi-agent traces, concurrency 8; 291 requests. \\
\texttt{weka-\allowbreak 20m-\allowbreak c24}
  & weka
  & 20.0
  & 8{,}331
  & Same corpus and scenario at concurrency 24; 333 requests, 11 client timeouts (3.3\% error rate). \\
\texttt{weka-\allowbreak 1h-\allowbreak c24}
  & weka
  & 60.0
  & 8{,}332
  & Same corpus at concurrency 24 over one hour; 822 requests, 40 timeouts. Long enough to outlive the metrics-window retention. \\
\addlinespace
\texttt{exgentic-\allowbreak 2m-\allowbreak c10}
  & exgentic-\allowbreak traces
  & 2.0
  & 8{,}738
  & Inference Perf non-streaming replay of \texttt{Exgentic/\allowbreak agent-\allowbreak llm-\allowbreak traces}, 10 concurrent sessions; 49 requests. First coverage window after engine start, so it also captures lazy imports. \\
\texttt{exgentic-\allowbreak 20m-\allowbreak c10}
  & exgentic-\allowbreak traces
  & 20.0
  & 8{,}678
  & Identical configuration over 20\,min; 393 requests, 2 failures. \\
\texttt{exgentic-\allowbreak 1h-\allowbreak c20}
  & exgentic-\allowbreak traces
  & 60.0
  & 8{,}811
  & Full trace corpus (no subset cap) at concurrency 20; 1{,}309 requests, 24 failures. Batches large enough to engage the fused expert router and the draft model's own forward pass. \\
\addlinespace
\texttt{01-bare-\allowbreak no-tools}
  & synthetic
  & 0.3
  & 7{,}282
  & Synthetic agentic baseline: 5 sessions, one turn each, no tools advertised, no fan-out. \\
\texttt{02-multiturn-\allowbreak no-tools}
  & synthetic
  & 0.5
  & 7{,}292
  & 5 sessions of 3 turns with growing context, still no tools; 98.2\% KV hit. \\
\texttt{03-single-agent-\allowbreak tool-loop}
  & synthetic
  & 2.2
  & 8{,}495
  & 5 sessions, tool-loop depth 2 over an 8-tool catalog. \\
\texttt{04-long-\allowbreak tool-loop}
  & synthetic
  & 5.2
  & 8{,}543
  & Tool-loop depth 5, exercising argument variety across a longer loop. \\
\texttt{05-multiturn-\allowbreak with-tools}
  & synthetic
  & 3.0
  & 7{,}978
  & 4 sessions of 3 turns, tool-loop depth 1, 6-tool catalog. \\
\texttt{06-parallel-\allowbreak tool-calls}
  & synthetic
  & 4.9
  & 8{,}516
  & Tool-loop depth 2 emitting 3 parallel tool calls per step; 70.8\% KV hit. \\
\texttt{07-chat-with-\allowbreak tools-varying}
  & synthetic
  & 7.2
  & 8{,}531
  & 4 sessions of 4 turns with tool-loop depth drawn uniformly from $[0,3]$, so depth varies per round. \\
\texttt{08-orchestrator-\allowbreak fanout}
  & synthetic
  & 4.9
  & 8{,}546
  & Orchestrator spawning two non-recursive children, each running a 3-turn tool loop. Degraded: 3 request failures, 1 of 4 sessions completed, 33 of 60 events cancelled. \\
\texttt{09-recursive-\allowbreak fanout}
  & synthetic
  & 2.6
  & 8{,}602
  & Recursive fan-out to depth 2 (root, children, grandchildren) over identical 200-token prefixes; 92 requests at 96.3\% KV hit. \\
\texttt{10-big-\allowbreak catalog}
  & synthetic
  & 1.2
  & 8{,}534
  & One turn per session but a 200-tool catalog per agent; 8 requests carrying 303{,}569 prompt tokens at 50.7\% KV hit. \\
\texttt{11-mixed-\allowbreak everything}
  & synthetic
  & 25.7
  & 8{,}678
  & 10 sessions blending every dimension: 1--3 turns, 40\% fan-out to depth 2, tool-loop depth 0--2, lognormal turn sizes; 373 requests. \\
\texttt{12-realistic-\allowbreak scale}
  & synthetic
  & 66.1
  & 8{,}730
  & Production-shaped: 20 sessions, 30\% fan-out to depth 2, tool-loop depth 1--4, the real 487-tool catalog, lognormal inputs averaging 110{,}000 tokens; 724 requests, 171{,}529{,}142 prompt tokens. \\
\texttt{13-high-\allowbreak turn-count}
  & synthetic
  & 16.7
  & 8{,}530
  & Long-conversation tail: 2 sessions of 32 turns, transcript growing each round; 97.8\% KV hit. \\
\texttt{14-high-tool-\allowbreak loop-depth}
  & synthetic
  & 13.1
  & 8{,}561
  & Long-tool-loop tail: tool-loop depth 32, i.e.\ 33 calls within a single turn; 66 requests. \\
\texttt{15-high-turns-and-\allowbreak tool-loop-\allowbreak distribution}
  & synthetic
  & 85.2
  & 8{,}592
  & 30 sessions with turns and tool-loop depth both drawn lognormally, giving many short sessions and a few very long ones; 725 requests, 171 of 896 events cancelled. \\
\texttt{16-budget-\allowbreak cap}
  & synthetic
  & 1.7
  & 8{,}541
  & Fully recursive fan-out to depth 3 deliberately halted by a budget cap---an expected stop condition, not an error; 40 requests. \\
\midrule
\textbf{Union} & & \textbf{559.4} & \textbf{9{,}778} & \textbf{24 runs across 5 families; 5{,}443 statements covered by every family} \\
\bottomrule
\end{tabular}
\end{table*}

\begin{table*}[tp]
\centering
\footnotesize
\caption{Statement coverage by static reachability class, over the union of
all 24 runs, including server startup. Reachability is computed on the
code as deployed in the serving image, so every measured file is classified.
The three unreachable classes act as a control: a statement executed in one would
falsify the partition, and no measurement window executes any.}
\label{tab:cumulative-coverage}
\begin{tabular}{@{}p{0.29\columnwidth}rrr@{}}
\toprule
\textbf{Class} & \textbf{Covered} & \textbf{Statements} & \textbf{Coverage} \\
\midrule
Reachable here & 74{,}562 & 239{,}319 & 31.16\% \\
\midrule
Features not enabled & 0 & 66{,}335 & 0\% \\
Other architectures & 0 & 109{,}926 & 0\% \\
Other accelerator & 0 & 4{,}091 & 0\% \\
\midrule
\textbf{Total} & \textbf{74{,}562} & \textbf{419{,}671} & \textbf{17.77\%} \\
\bottomrule
\end{tabular}
\end{table*}

\begin{table*}[tp]
\centering
\small
\caption{Coverage by test family. Coverage is the union of the statements the
family's runs executed in their own isolated windows; unique is the statements
in that union which no \emph{other} family reached. Families are ordered by unique
contribution. Across all five families, 9{,}778 distinct statements were
covered, of which 5{,}443 were reached by every family. Separately, 437
statements were reached by exactly one family; those are what the unique column
counts. The remaining statements were
reached by more than one family but not all five. Startup is excluded here,
unlike Table~\ref{tab:cumulative-coverage}, so these figures are request-driven
execution only. Unique coverage is a property of this particular partition into
five families: a statement two families reach counts as unique to neither. Statements in
the structured-output subsystem are excluded from the unique column, because
they are reached at engine initialization rather than by any family's traffic
(Section~\ref{app:coverage-method}).}
\label{tab:unique_coverage}
\begin{tabular}{@{}p{0.13\textwidth}rrrrp{0.46\textwidth}@{}}
\toprule
\textbf{Family} & \textbf{Runs} & \textbf{Duration} & \textbf{Coverage} & \textbf{Unique} & \textbf{What only this family reached} \\
\midrule
live & 1 & 136.8 & 8{,}777 & \textbf{331} &
The only family to use the \textbf{Anthropic messages API}: its request handler, router, protocol models, and the token scanner that drives incremental parsing on that path. \\
weka & 3 & 100.0 & 8{,}431 & 54 &
The only \textbf{streaming} family, which shifts tool-call parsing to different partial-token positions. In the \textbf{KV offload scheduler}, the only family to reconcile \emph{deferred fences}: blocks still in flight to CPU as their requests finish, reconciled when those blocks come back for reuse. \\
exgentic-\allowbreak traces & 3 & 82.0 & 8{,}967 & 27 &
The only family to enter the \textbf{short-context branch of sparse attention}, taken when a request offers fewer index candidates than the top-$k$ width so the indices are filled directly rather than selected. Also tool-call parser states reached only by the recorded trace mix: parser-engine transitions, the token scanner's anchor-rescan path, and the tool-argument JSON wrapper's non-finite-value fallback. \\
synthetic & 16 & 240.5 & 8{,}795 & 5 &
Two tool-argument coercion branches, one parser fallback, and the workspace-area cutoff in \textbf{sparse-SWA chunk planning}, reached only by this family's very long prompts. \\
smoke & 1 & 0.1 & 5{,}748 & 20 &
\textbf{Greedy sampling}: at temperature 0 the sampling parameters take the zero-temperature branch that every other workload skips. Also the profiling-scope and first-request metrics paths that only a cold single request reaches. \\
\midrule
\textbf{Union} & \textbf{24} & \textbf{559.4} & \textbf{9{,}778} & \textbf{437} & \\
\bottomrule
\end{tabular}
\end{table*}

\begin{table*}[tp]
\centering
\scriptsize
\caption{Hard-to-reach scheduling, caching and KV-offload states, and which
workloads drove the serving stack into them. These are engine-internal
transitions---queue backpressure, prefix-cache collisions, eviction, in-flight
transfer reconciliation---rather than API surface, and each requires a
particular kind of pressure rather than a particular request format. Each cell
counts how many statements of that state's code the family executed, out of the
total in the \textbf{Stmts} column; a dash means the family never entered that
code at all. Italics in the last column describe the state itself, and the
sentence after says what a workload must do to provoke it. The states are
identified by the statements that no single-request run reaches; line numbers are
from the deployed image (vLLM \texttt{v0.31.0}). The \texttt{smoke} column is
a dash throughout: a correct single request, served end to end, enters none of
these paths.}
\label{tab:load-states}
\begin{tabular}{@{}p{0.29\textwidth}c@{\hspace{1.2em}}ccccc p{0.33\textwidth}@{}}
\toprule
\textbf{State, and the code that enters it} & \rotatebox{90}{\textbf{Stmts}} & \rotatebox{90}{\textbf{smoke}} & \rotatebox{90}{\textbf{synthetic}} & \rotatebox{90}{\textbf{exgentic}} & \rotatebox{90}{\textbf{weka}} & \rotatebox{90}{\textbf{live}} & \textbf{What the state is (\emph{italics}), and what it takes to reach it} \\
\midrule

\textbf{Queue backpressure: defer a waiting request that already holds KV blocks} \newline
\texttt{vllm/v1/core/sched/\allowbreak scheduler.py} 878--883, 1345 \newline
{\scriptsize\texttt{req = from\_queue.pop\_request()}} \newline
{\scriptsize\texttt{self.deferred\_waiting.add(req)}} \newline
{\scriptsize\texttt{if self.\_holds\_kv\_blocks(req):}} \newline
{\scriptsize\texttt{~~step\_skipped\_kv\_holding.appendleft(req)}}
  & 7 & -- & \textbf{7} & \textbf{7} & -- & --
  & \emph{The scheduler cannot start every queued request in one step, and one it must postpone is already holding GPU cache blocks from a previous turn. It has to set that request aside without losing track of the blocks.} Reaching it means offering more concurrent sessions than the token budget admits. Only the two generated and replayed agent families sustained enough queued work; the hour-long WEKA replays at concurrency 24 never filled the budget this way. \\
\addlinespace

\textbf{Prefix-cache hash collision: one block hash mapping to several blocks} \newline
\texttt{vllm/v1/core/\allowbreak block\_pool.py} 68--69, 91--97 \newline
{\scriptsize\texttt{elif isinstance(blocks, KVCacheBlock):}} \newline
{\scriptsize\texttt{~~self.\_cache[key] = \{blocks.block\_id:}} \newline
{\scriptsize\texttt{~~~~blocks, block.block\_id: block\}}} \newline
{\scriptsize\texttt{if isinstance(blocks, dict):}} \newline
{\scriptsize\texttt{~~return next(iter(blocks.values()))}}
  & 6 & -- & \textbf{6} & \textbf{6} & 2 & --
  & \emph{Two different cache blocks turn out to have the same content hash, so the prefix index can no longer store one block per key and must keep a set of them.} Reaching it means sending many sessions whose prompts share an identical prefix. Recursive agent fan-out over a common system prompt is what produces this. \\
\addlinespace

\textbf{LRU heap compaction while evicting from the CPU tier} \newline
\texttt{vllm/v1/kv\_offload/cpu/\allowbreak policies/lru.py} 55--70, 168 \newline
{\scriptsize\texttt{if len(self.\_heap) < max(64,}} \newline
{\scriptsize\texttt{~~~~2 * len(self.\_evictable)): return}} \newline
{\scriptsize\texttt{self.\_heap = [(self.\_ranks[k], k)}} \newline
{\scriptsize\texttt{~~~~for k in self.\_evictable]}} \newline
{\scriptsize\texttt{heapq.heapify(self.\_heap)}}
  & 9 & -- & 7 & 7 & \textbf{9} & 7
  & \emph{The CPU tier is full and must evict. The heap that tracks least-recently-used blocks has accumulated so many stale entries that it is cheaper to rebuild it from scratch than keep sifting it.} Reaching it means filling the CPU tier and then continuing to touch blocks already in it. Every sustained family reaches this; only \texttt{weka} drives the recency-update path that rebuilds the heap from the evictable set. \\
\addlinespace

\textbf{Deferred KV fence reconciled when a block is reused} \newline
\texttt{vllm/distributed/kv\_transfer/\allowbreak kv\_connector/v1/offloading/\allowbreak scheduler.py} 654--658, 1971--1973 \newline
{\scriptsize\texttt{for bid in job\_status.}} \newline
{\scriptsize\texttt{~~~~deferred\_fence\_block\_ids or ():}} \newline
{\scriptsize\texttt{~~self.\_block\_id\_to\_pending\_jobs}} \newline
{\scriptsize\texttt{~~~~.setdefault(bid, set()).add(job\_id)}}
  & 8 & -- & -- & -- & \textbf{8} & --
  & \emph{A request finished, but its cache blocks were still being copied to CPU. The server cannot free them yet, so it records a marker and settles the copy later, when something else asks for those same blocks.} Reaching it means completing responses while transfers are still in flight. Reached only by streaming replay, where responses complete while transfers are still in flight. \\
\addlinespace

\textbf{Bulk cancellation across the waiting, KV-holding and deferred queues} \newline
\texttt{vllm/v1/core/sched/\allowbreak scheduler.py} 2608--2630 \newline
{\scriptsize\texttt{if request.status == RequestStatus.RUNNING:}} \newline
{\scriptsize\texttt{~~running\_requests\_to\_remove.add(request)}} \newline
{\scriptsize\texttt{self.waiting.remove\_requests(}} \newline
{\scriptsize\texttt{~~~~waiting\_requests\_to\_remove)}}
  & 6 & -- & -- & -- & \textbf{6} & \textbf{6}
  & \emph{Requests are dropped while still waiting rather than while running, so the scheduler has to pull them out of several different queues at once instead of retiring them normally.} Reaching it means clients going away before they are served. Client timeouts under replay and terminated agent sessions both produce it; no generated workload did. \\
\addlinespace

\textbf{Speculative draft tokens trimmed and padded after scheduling} \newline
\texttt{vllm/v1/core/sched/\allowbreak scheduler.py} 2515--2547, 2917 \newline
{\scriptsize\texttt{del spec\_token\_ids[orig\_num\_spec\_tokens:]}} \newline
{\scriptsize\texttt{num\_invalid = orig\_num\_spec\_tokens}} \newline
{\scriptsize\texttt{~~~~- len(spec\_token\_ids)}} \newline
{\scriptsize\texttt{spec\_token\_ids.extend([-1] * num\_invalid)}}
  & 10 & -- & \textbf{10} & \textbf{10} & -- & --
  & \emph{Speculative decoding guesses several tokens ahead, but the scheduler admitted fewer tokens for this request than the drafter guessed, so the surplus guesses are cut and the gap padded out.} Reaching it means batches large enough that drafting and admission disagree. Requires concurrent decode and chunked prefill in the same step. \\
\addlinespace

\bottomrule
\end{tabular}
\end{table*}

\subsection{Closing the KV-restore coverage hole with a targeted synthetic
workload}
\label{app:coverage-kvrestore}

The coverage analysis of Section~\ref{sec:vllm_coverage} left one gap that no
workload in Table~\ref{tab:coverage-workloads} closed: GPU-to-CPU offloading
was entered by every sustained family, but the reverse CPU-to-GPU
\emph{restore} path was never executed. This appendix reports the follow-up
experiment that closed it, because it is the clearest demonstration we have of
what a parametrically controlled synthetic workload buys that neither replay
nor live agents did.

\paragraph{The target.} The restore path is entered through
\texttt{submit\_load} in
\texttt{vllm/v1/kv\_offload/cpu/gpu\_worker.py}, which forwards to
\texttt{\_load\_handler.transfer\_async}.

\paragraph{Why no earlier workload reached it.} A block is promoted back to
GPU only if it was evicted from GPU first: \texttt{lookup()} in
\texttt{vllm/v1/kv\_offload/tiering/manager.py} returns early on a primary
cache hit and never consults the CPU tier. Reaching the restore path therefore
requires a conjunction of two properties that no recorded corpus in our set
happened to satisfy at once: a working set that \emph{exceeds} the GPU KV pool,
so that eviction occurs, and revisits to an evicted prefix that are spaced far
enough apart in wall-clock time that the prefix is evicted before it is asked
for again. The WEKA and Exgentic replays and the CyberGym live run all filled
and evicted from the CPU tier---Table~\ref{tab:load-states} shows them
reaching the LRU compaction path---but their traffic never produced the second
property.

\paragraph{Workload design.} We used the \texttt{shared\_prefix} synthetic data
generator in Inference Perf, which constructs $G$ groups of prompts sharing a
per-group system prompt of a configured length.  Figure~\ref{fig:kvrestore-config} gives the configuration.

The sizing argument is explicit. The deployment reports a GPU KV pool of
9{,}580{,}564 tokens (about 149{,}696 blocks at \texttt{--block-size 64}). We
chose $200$ groups $\times$ $65{,}536$ tokens $=13{,}107{,}200$ tokens,
or $1.37\times$ the pool: above $1.0\times$, so eviction is guaranteed, and far
enough above it that the approximate pool figure read from the server log
cannot push the true ratio below 1. The 512\,GiB CPU tier is large enough that
this working set does not saturate it.

\begin{figure}[t]
\centering
\begin{lstlisting}[style=trace]
data:
  type: shared_prefix
  shared_prefix:
    num_groups: 200          # x 65,536 tok = 1.37x GPU KV pool
    num_prompts_per_group: 3
    system_prompt_len: 65536
    question_len: 256
    output_len: 128
    seed: 20261029
load:
  type: constant
  stages:
    - rate: 1                # << measured ~2.5 req/s capacity
      duration: 900
  num_workers: 4
  worker_max_concurrency: 4  # hard cap: no backlog can form
  request_timeout: 600
  stage_teardown_grace_seconds: 240.0
\end{lstlisting}
\caption{The workload configuration that closed the KV-restore coverage hole
(Inference Perf, \texttt{shared\_prefix} generator). The two parameters that
matter are the working set, set by \texttt{num\_groups} $\times$
\texttt{system\_prompt\_len} to $1.37\times$ the GPU KV pool so that eviction
is guaranteed, and the offered \texttt{rate}, held well below
measured capacity so that revisits to a prefix are spaced far enough apart for
it to be evicted first.}
\label{fig:kvrestore-config}
\end{figure}

\paragraph{Result.} Table~\ref{tab:kvrestore} reports the outcome on two
independent signals. The pod had restarted before this run, so the server
counters start from a true zero baseline.

\begin{table}[t]
\centering
\footnotesize
\caption{The KV-restore run. Server metrics are
\texttt{vllm:kv\_offload\_*} counters read from the server before and after the
run, from a true zero baseline. Coverage is from the run's own isolated window:
line 895 is the entire executable body of \texttt{submit\_load}, and
\emph{request-driven} means it was executed inside that window rather than at
import. Line 891 is the \texttt{def}, executed at import by any run that loads
the module, which is why it cannot serve as the signal.}
\label{tab:kvrestore}
\setlength{\tabcolsep}{4pt}
\begin{tabular}{@{}lrr@{}}
\toprule
\textbf{Server metric (CPU\,$\to$\,GPU)} & \textbf{before} & \textbf{after} \\
\midrule
\texttt{kv\_offload\_total\_bytes\_total} & 0 & 108.1\,GB \\
\texttt{kv\_offload\_size\_count} & 0 & 2{,}816 ops \\
\texttt{kv\_offload\_total\_time\_total} & 0 & 3.33\,s \\
\addlinespace
GPU\,$\to$\,CPU bytes (for reference) & 0 & 193.8\,GB \\
\midrule
\textbf{Coverage, \texttt{gpu\_worker.py}} & \multicolumn{2}{c}{} \\
\midrule
line 891, \texttt{def} & \multicolumn{2}{r}{covered at import} \\
line 894, docstring & \multicolumn{2}{r}{not covered} \\
\textbf{line 895, body} & \multicolumn{2}{r}{\textbf{covered, request-driven}} \\
\bottomrule
\end{tabular}
\end{table}

\begin{center}
\footnotesize
\begin{tabular}{@{}lr@{}}
\toprule
\textbf{File (under \texttt{vllm/})} & \textbf{+stmts} \\
\midrule
\texttt{distributed/.../offloading/scheduler.py} & 82 \\
\texttt{v1/kv\_offload/cpu/manager.py} & 25 \\
\texttt{distributed/.../offloading/worker.py} & 6 \\
\texttt{distributed/.../offloading/metrics.py} & 3 \\
\texttt{v1/kv\_offload/tiering/manager.py} & 3 \\
\texttt{distributed/.../offloading/events.py} & 2 \\
\texttt{v1/kv\_offload/cpu/gpu\_worker.py} & 1 \\
\texttt{v1/kv\_offload/cpu/policies/lru.py} & 1 \\
\texttt{v1/kv\_offload/tiering/metrics.py} & 1 \\
\midrule
\textbf{Offload/tiering subsystem} & \textbf{124} \\
\bottomrule
\end{tabular}
\end{center}

\subsection{Limitations in detail}
\label{app:coverage-limits}

The coverage results of Section~\ref{sec:vllm_coverage} highlight the
complementary value of different testing approaches by identifying code
uniquely exercised by specific workloads. However, given the limitations
below, similar statement coverage should not be immediately interpreted as evidence
that tests are redundant: two harnesses may execute the same statements while
checking different properties of the result.

\begin{itemize}
  \item \textbf{Python only.} Only the interpreted parts of vLLM (the API, Scheduler, Sampling, and KV-Offloader) are measured. Parts of vLLM, and specifically the
        attention and matrix-multiply kernels, are compiled C++, CUDA, or Rust and are not visible.

  \item \textbf{Execution is not correctness.} A covered line was executed,
        not shown to behave correctly; different approaches have different result-checking capabilities (e.g., \texttt{live} checks quality, while other approaches can only identify performance issues or crashes).

  \item \textbf{Single configuration.} Results are for one model on one
        hardware configuration of one vLLM version. We expect the broad
        ordering of approaches to transfer, but we have not tested this: with
        the \texttt{live} and \texttt{smoke} families represented by a single
        run each, the ranking rests on few observations, and the specific
        statement sets will certainly not transfer.

  \item \textbf{Unique coverage is sensitive to the family partition.} A
        statement reached by two families counts as unique to neither, so adding a
        workload can only reduce another family's unique count. The figures in
        Table~\ref{tab:unique_coverage} therefore describe this set of five
        families, not the approaches in general.

  \item \textbf{The reachable-statement count is an over-estimate.}
        Because reachability is resolved per file, the 239{,}319-statement
        denominator is an over-estimate and 31.16\% is correspondingly a floor. A
        module imported for one symbol counts as reachable even when most of its body
        cannot run (for example, the
        vendored CuTeDSL and FlashAttention kernel emitters: 27{,}096 statements
        stay reachable though most
        emit kernels that a sparse-MLA inference server never
        calls).
\end{itemize}

%% file: appendix_repr.tex
\section{Production Workload Representativeness}
\label{app:repr}

We experimented with typical agentic workloads on the GLM-5.3 deployment. We compared per-request token lengths
 against the public WEKA trace and the traces captured before deployment.

\emph{Data and confidentiality.} Traffic is subject to
confidentiality restrictions, so we report only aggregate statistics: the
per-request input and output length histograms exported by the serving stack
(vLLM Prometheus metrics, 20 buckets from 1 to $10^6$ tokens), which contain
token counts but no request content or session identifiers. The comparison is
therefore at bucket resolution. A session-level comparison, covering calls per
session, session duration, and arrival patterns, is out of the scope of this
paper.\\
\emph{Setup.} We use a 4-day traffic window of about 2.2M requests, and bin
every candidate workload into the same 20 buckets. The candidates are the
public WEKA trace~\citep{semianalysis_weka_060826}
(94,298 main-agent requests, with token counts as provided by the dataset) and
traces captured with Exgentic on CyberGym, DeepSWE, HWE-Bench, and HANDBOOK~\citep{panavas2026handbookmdbenchmarklongcontextagentic}
using the same model (1.6K--16.4K calls each). We report the
Kolmogorov--Smirnov (KS) distance between cumulative distributions (0 means
identical). A second, non-overlapping 4-day window yields nearly identical
workload distributions (KS 0.02--0.06) and the same conclusions.

\begin{figure*}[t]
\centering
\includegraphics[width=\textwidth]{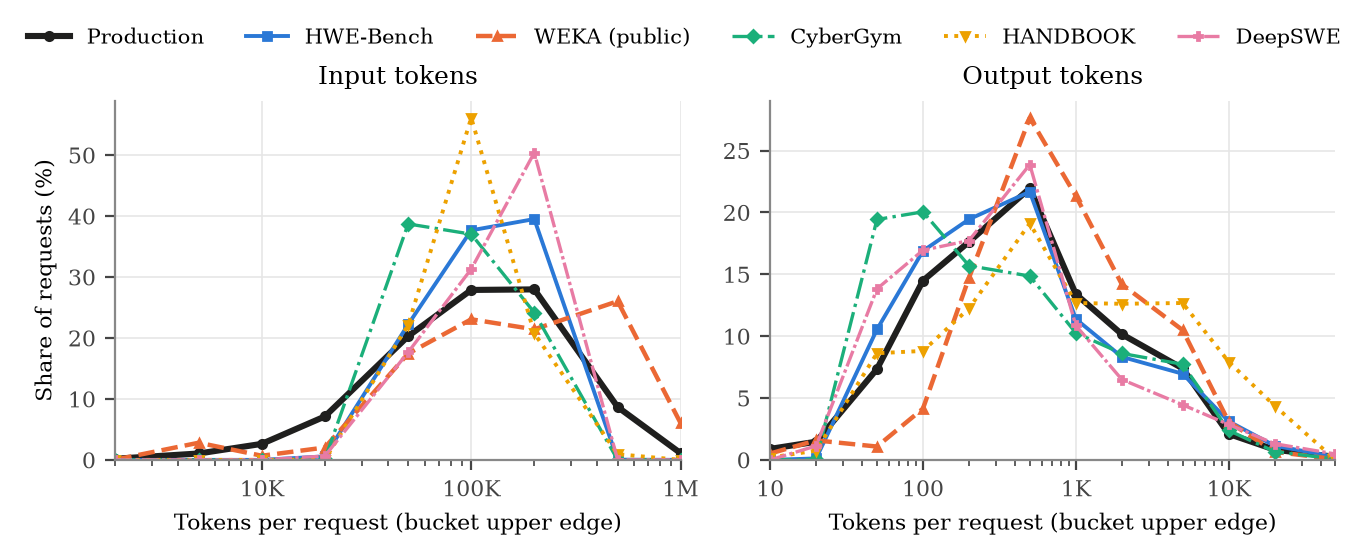}
\caption{Distribution of per-request input and output token lengths for
the GLM-5.3 typical agentic workload, the public WEKA trace, and the four traces captured on
the target model. Each point is the share of requests in one workload
histogram bucket, plotted at the bucket's upper edge.}
\label{fig:repr}
\end{figure*}

\emph{Results.} The typical agentic workload is long-context with short outputs: the
median request carries 50K--100K input tokens and produces 200--500 output
tokens. As Table~\ref{tab:repr} and Figure~\ref{fig:repr} show, WEKA is the
furthest from our workload in both input and output lengths. Its input
distribution is shifted toward longer prompts (median 100K--200K tokens, with
about a third of requests above 200K), and its output distribution has fewer
short responses than our workload. The captured traces all fall in
the workload's input range, but their shapes differ: CyberGym produces shorter
outputs, HANDBOOK has a long output tail, and DeepSWE concentrates its inputs
at 100K--200K. HWE-Bench is the closest overall, with an output distribution
that nearly matches the workload's distribution (KS 0.04, about $5\times$ closer than WEKA).
No single benchmark reproduces the typical agentic workload on its own, which supports
capturing several use-case benchmarks rather than relying on a single trace.

\begin{table}[t]
\caption{KS distance between each workload and the GLM-5.3 typical agentic workload,
per request (lower is closer).}
\label{tab:repr}
\centering
\small
\begin{tabular}{lcc}
\toprule
Workload & Input tokens & Output tokens \\
\midrule
HWE-Bench  & \textbf{0.135} & \textbf{0.044} \\
CyberGym   & 0.140 & 0.133 \\
HANDBOOK   & 0.160 & 0.164 \\
DeepSWE     & 0.161 & 0.081 \\
WEKA         & 0.223 & 0.212 \\
\bottomrule
\end{tabular}
\end{table}

